\documentclass{statsoc}  % not working properly -- borders wrong
\usepackage{url}            % simple URL typesetting
\usepackage{booktabs}       % professional-quality tables
\usepackage{amssymb}
\usepackage{amsfonts}       % blackboard math symbols
\usepackage{nicefrac}       % compact symbols for 1/2, etc.
\usepackage{natbib}
\setcitestyle{authoryear}

\usepackage{graphicx}
\usepackage{color}
\usepackage{amsmath}
\usepackage{enumitem}
\usepackage{bm}
\usepackage{xcolor}
\newenvironment{itemize*}{
\begin{itemize}
\setlength{\parskip}{0em}
\setlength{\topskip}{0em}
}
{\end{itemize}}

\newenvironment{enumerate*}{
\begin{enumerate}
\setlength{\parskip}{0em}
\setlength{\topparskip}{0em}
}
{\end{enumerate}}

\definecolor{mygray}{gray}{.30}
\newtheorem{prop}{Proposition}
\newtheorem{thm}[prop]{Theorem}
\newtheorem{assumption}{Assumption}

\newtheorem{itheorem}{Stability Theorem}  %newtheorem* doesn't work for JRSS

\newcommand{\beq}{\begin{equation}}
\newcommand{\eeq}{\end{equation}}
\newcommand{\beqa}{\begin{eqnarray}}
\newcommand{\eeqa}{\end{eqnarray}}
\newcommand{\beqas}{\begin{eqnarray*}}
\newcommand{\eeqas}{\end{eqnarray*}}

\newcommand{\bit}{\begin{itemize}}
\newcommand{\eit}{\end{itemize}}
\newcommand{\bits}{\begin{itemize*}}
\newcommand{\eits}{\end{itemize*}}
\newcommand{\benum}{\begin{enumerate}}
\newcommand{\eenum}{\end{enumerate}}
\newcommand{\benums}{\begin{enumerate*}}
\newcommand{\eenums}{\end{enumerate*}}
\newcommand{\comment}[1]{}

\newcommand{\probe}{\text{(SS\kmeans)}}
\newcommand{\probncut}{\text{(SS\ncut)}}
\newcommand{\probstar}{\ensuremath{({\large *})}} % the generic stability result

\newcommand{\todo}[1]{}%{\textcolor{red}{\em TODO: {#1}}}
\newcommand{\mmp}[1]{}
\newcommand{\hanyuz}[1]{}
\newcommand{\techreport}[1]{}

\newcommand{\mf}[1]{}%\textcolor{green}{MF: #1}}

\newcommand{\mydef}[1]{{\em #1}}
\newcommand{\myemph}[1]{{\em #1}}

\newcommand{\mfcv}{DFCV}

\newcommand{\bfone}{{\mathbf 1}}
\newcommand{\rrr}{{\mathbb R}}

\newcommand{\clust}{{\cal C}}
\newcommand{\C}{{\cal C}}  % to remove, but from where

\newcommand{\X}{{\mathcal X}} % relaxed space of clustering
\newcommand{\T}{{\mathcal T}} % technical conditions
\newcommand{\clustal}{\mathbf{{C}}_K}

\newcommand{\norm}[1]{\lvert\lvert{#1}\lvert\lvert}
\newcommand{\tilX}{\tilde{X}}

\newcommand{\loss}{\ensuremath{\operatorname{Loss}}}
\newcommand{\kmeans}{\ensuremath{_{\rm Km}}}
\newcommand{\lp}{\ensuremath{_{\rm LP}}}
\newcommand{\lossk}{\loss\kmeans}

\newcommand{\ncut}{\ensuremath{_{\rm NCut}}}
\newcommand{\llle}[1]{_{\leq{#1}}}
\newcommand{\llll}{\llle{l}}

\newcommand{\opt}{^{{\rm opt}}}  % refers to the optimal clustering, while * refers to the optimal solution of an optimization problem, usually a relaxation

\newcommand{\dataset}{{\cal D}}
\newcommand{\Pp}{{\cal P}}

\newcommand{\diag}{\operatorname{diag}}

\newcommand{\trace}{\operatorname{trace}}
\newcommand{\distance}{d^{EM}}

\newcommand{\epsi}{\varepsilon}

\newcommand{\pmin}{p_{\rm min}}
\newcommand{\pmax}{p_{\rm max}}

\newlength{\picwi}

\title{Distribution free optimality intervals for clustering}
\author{Marina Meil\u{a}}
\address{Department of Statistics, University of Washington, Seattle, USA}
\email{mmp@stat.washington.edu}
\author[Marina Meil\u{a} and Hanyu Zhang]{Hanyu Zhang}
\address{Department of Statistics, University of Washington, Seattle, USA}

\begin{document}

\maketitle

\begin{abstract}
We address the problem of validating the ouput of clustering
algorithms. Given data $\dataset$ and a partition $\clust$ of these
data into $K$ clusters, when can we say that the clusters obtained are correct or meaningful for
the data? This paper introduces a paradigm in which a clustering
$\clust$ is considered meaningful if it is {\em good} with respect to a
 {\em loss function} such as the K-means distortion, and
{\em stable}, i.e. the only good clustering up to small
perturbations. Furthermore, we present a generic method to obtain
post-inference guarantees of near-optimality and stability for a
clustering $\clust$. The
method can be instantiated for a variety of clustering criteria (also called loss functions)
for which convex relaxations exist. Obtaining the guarantees amounts
to solving a convex optimization problem.  We demonstrate the practical
relevance of this method by obtaining guarantees for the K-means and the Normalized Cut clustering criteria on
realistic data sets. We also prove that asymptotic instability implies finite sample instability w.h.p., allowing inferences about the population clusterability from a sample. The guarantees do not depend on any
distributional assumptions, but they depend on the data set $\dataset$ admitting a stable clustering.
%and in particular the guarantees exist only when the data is {\em  clusterable}.
\end{abstract}

\section{Introduction}
\label{sec:introduction}

We are concerned with the problem of finding structure in data by clustering. This is an old problem, yet some of the most interesting advances in its theoretical understanding are recent. Namely, while empirical evidence has been suggesting that when the data are ``well clustered'', the cluster structure is easy to find, proving this in the non-asymptotic regime is an area of current progress.  For instance, it was shown that finding an almost optimal clustering for the K-means and K-medians cost can be done by efficient algorithms when clusters are ``well separated'' \citet{Aswathi+Ward:14}, or when the optimal clustering is ``resilient'' \cite{balcanL:16}. These results are significant because they promise computationally efficient inference in the special case when the data are ``clusterable'', even though most clustering loss functions induce hard optimization problems in the worst case.
\hanyuz{ For this paragraph, now given that the reviewers are not very familiar with the context of clustering, this paragraph might be too vague and hard to follow for them. I suggest starting with a more fundamental introduction to what is clustering. Avoid these new advances in TCS (only mention them in the related work?) We could say like this: clustering is the question of ..... We have developed many algorithms for clustering, among them, loss based clustering is a very typical and widely used algorithm. For example, K-means...., Then we can provide a technical definition of K-means clustering in a statistical framework (like in Pollard's paper of K-means consistency) And say that K-means has some good properties including consistency and asymptotic normality. However it would be hard to evaluate how good is the result of K-means clustering.  And then we introduce the notion of stability. }

The present paper addresses the post-inference aspect of clustering. We propose a framework for providing guarantees for a given clustering $\clust$ of a given set of points, without making untestable assumptions about the data generating process. In the simplest terms, the question we address is: can a user tell, with no prior knowledge, if the clustering $\clust$ returned by a clustering algorithm is meaningful? or correct? or optimal?

\hanyuz{Maybe for a statistical journal this paragraph is vague and too colloquial? I feel that they would prefer a formal mathematical definition.}

\comment{
While the above algorithms provably exploit clusterability of the data under various definitions, there is relatively little work on verifying that these assumptions hold for the data at hand. Without this validation step, the theoretical guarantees promised by this body of work cannot be used in practice. }

This is the fundamental problem of cluster validation, and we call it
{\em Distribution Free Cluster Validation (\mfcv)}. We offer a general
data driven \mfcv~paradigm, in the case of {\em loss-based}
clustering.  In this framework, for a given number of clusters $K$,
the best clustering of the data is the one that minimizes a {\em loss
  function} $\loss(\dataset,\clust)$. This framework includes K-means,
K-medians, graph partitioning, etc. Note that these loss functions as
well as most clustering losses in use are NP-hard to optimize\mmp{find a ref about clustering here. probably one for each type of clustering}~
\cite{GareyJohnson:79}. In particular, this means that it is in general not possible
to prove that the optimum has been found \cite{GareyJohnson:79}. Thus, the
\mfcv~problem needs to be carefully framed before solving it is
attempted.

This paper will show that it is possible to obtain a
guarantee of ``correctness'' for a clustering precisely in those cases
when the data at hand admits a good clustering; in other words, when
the data is ``clusterable''.

\subsection{\bf Framing the problem.}
{\bf The loss as goodness of fit} How, thus, do we decide if $\clust$ is a
``meaningful'' clustering for data $\dataset$? First, $\clust$ must be
``good'' with respect to whatever definition of clustering we are
working with. This definition is embodied by the {\em loss function} $\loss$.
For example, in the case of K-means clustering, the loss function $\loss\kmeans$ is the quadratic distortion defined in equation \eqref{eq:loss0}, and in the case of graph partitioning by Normalized Cuts \cite{MShi:aistats01} the loss function $\loss\ncut$ is defined in equation \eqref{eq:NCut}.
Only a clustering $\clust$ that has low loss can be considered
meaningful; indeed, if the value of $\loss(\clust)$ was much larger
than the loss for other ways of grouping the data, $\clust$ could not
be a good or meaningful way to partition the data according to the user's definition of
clustering. Hence, the proposed \mfcv~framework the $\loss$ is a precise way to express what
a user means by ``cluster'' from the multitude of definitions
possible.

\hanyuz{Also as I am reading this, I tried to find if there could be anyway of mathematically defining what is "clusterability".  I found the paper by Margareta and Shai Ben-David about Clusterability: A theoretical Study, where they reviewed some clusterability criteria, e.g. center pertubation clusterability. Where this concept is also connected with the loss and could be very similar to our notion of stability. I feel that it is necessary to understand what's the connection between their clusterability and our stability.}

{\bf Stability (uniqueness)} Second, $\clust$ must be the {\em only}
``good'' clustering supported by the data $\dataset$, up to small
variations. This property is called \mydef{stability}. In the present paper  results of the following form will be obtained.

\begin{itheorem}[$*$]
Given a clustering $\clust$ of data set $\dataset$, a loss function $\loss()$, and technical conditions $\T$, there is an $\epsilon$ such that $\distance(\clust,\clust')\leq \epsilon$ whenever $\loss(\dataset,\clust')\leq \loss(\dataset,\clust)$.
\end{itheorem}
In the above $\distance(\clust,\clust')$ is the widely used {\em
  earth mover's distance (EM)} distance between partitions of a set of
$n$ objects, formally defined in Section \ref{sec:defs}. Theorem
$*$ would be trivial if $\epsilon$ was arbitrarily large; hence, we
shall always require that $\epsilon$ be small, for instance, smaller
than the relative size of the smallest cluster in $\clust$.

\begin{figure}
\setlength{\picwi}{0.3\textwidth}
\begin{tabular}{ccc}
\hspace{-2em}{\small Good, stable $\clust$} 
&
{\small Bad  $\clust'$} 
&
{\small Unstable  $\clust''$}
\\
\hspace{-1em}
\includegraphics[width=\picwi]{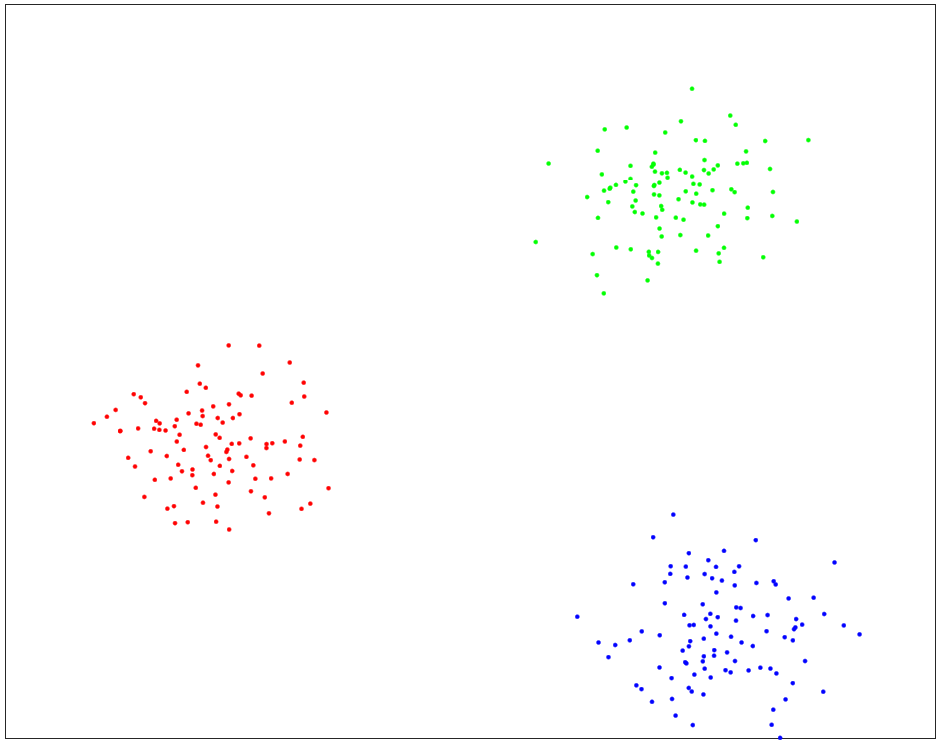}
&
\includegraphics[width=\picwi]{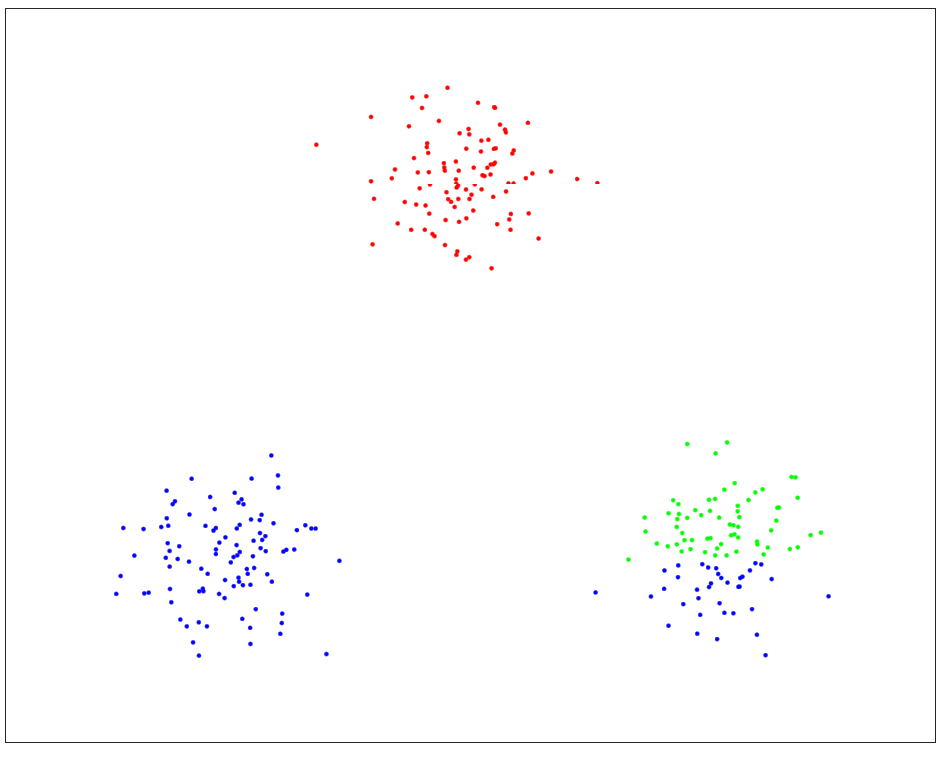}
&
%\hspace{-2em}
\includegraphics[width=1.05\picwi]{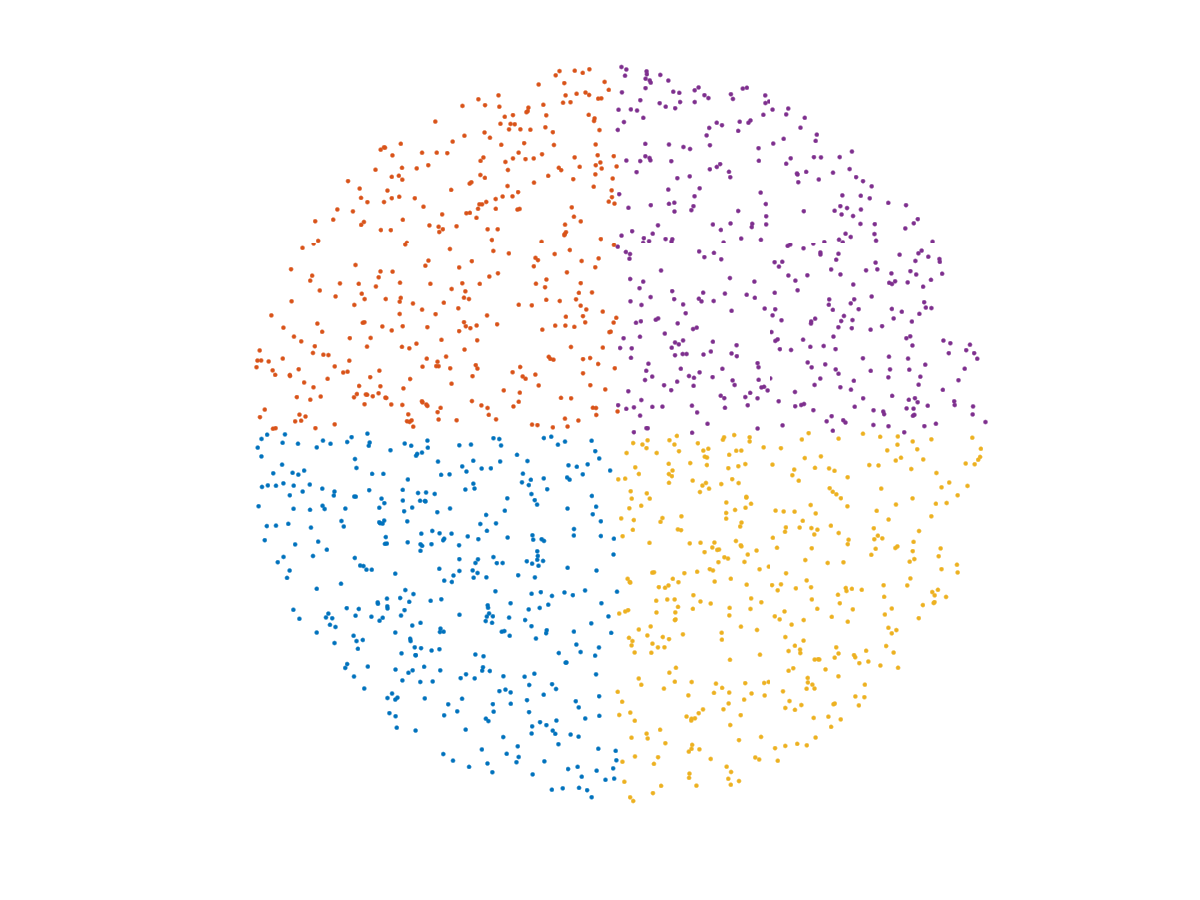}
\\
OI $\epsilon=10^{-4}$ & no guarantee & no guarantee 
\\
\end{tabular}
\caption{\small \label{fig:example-kmeans}
Left: a clusterable data set. The method described in Section \ref{sec:kmeans} guarantees than all clusterings of this data set that are at least as ``good'' as $\clust$ differ from it in no more $\epsilon=0.01\%$ of the points; since $\epsilon<1/n$, this guarantee also implies that $\clust$ is optimal w.r.t. \loss. Middle: the same data, with a clustering $\clust'$ which is neither good, nor stable. Right: a data set that is not clusterable. The clustering shown is nearly optimal, but it is not stable, as the data admits other clusterings with similar \loss, but very different from $\clust''$.
}
\end{figure}

A Stability Theorem states that any way to partition the data which is
very different from $\clust$ will result in higher cost. Hence, the
data supports only one way to be partitioned with low cost, and small
perturbations thereof. If a statement such as $(*)$ holds for a
clustering $\clust$, it means that $\clust$ captures structure
existing in the data, thus it is meaningful. It should also be
evident that it is not possible to obtain such guarantees in general;
they can only exist for specific data sets and clusterings, as illustrated in Figure \ref{fig:example-kmeans}. A clustering satisfying the Stability Theorem is called $\epsilon$-{\em stable} (or simply {\em stable}), and a data set that admits an $\epsilon$-stable clustering is said to be $\epsilon$-{\em clusterable} (or simply {\em clusterable}).

From Theorem $(*)$ it follows immediately that $C\opt$, the clustering minimizing \loss~on $\dataset$, together with the entire {\em sublevel set} $\{\clust',\,\loss(\clust')\leq \loss(\clust)\}$ is contained in the ball of radius $\epsilon$ centered at $\clust$. We call this ball an {\em optimality interval (OI)} for $\loss$ on $\dataset$. Somehow abusively, we will occasionally call $\epsilon$ itself an OI. 
    
It is important that the conditions of the Stability Theorem and the value of
$\epsilon$ above be \myemph{computable} from $\dataset$ and $\clust$. That
is, they must be tractable to evaluate, and must not contain undefined
constants. Figure \ref{fig:example-kmeans} illustrates this framework. 

To summarize, the Distribution Free Cluster Validation framework
starts with a loss function $\loss(\dataset,\clust)$ which serves both
as {clustering paradigm}, by defining implicitly the type of clusters
appropriate for the application, and as goodness of fit
measure. Supposing that it is possible to find a good clustering
$\clust$, the challenge is to verify that $\clust$ is stable without
enumerating all the other possible $K$-partitions of the data. Hence, the main
technical contribution of this work is to show how to obtain
conditions $\T$ that can be verified tractably. The key idea is to uses
convex tractable relaxations to the original \loss~minimization
problem.

After the definitions in Section \ref{sec:defs}, in Section \ref{sec:generic} we introduce the {\em Sublevel Set (SS)} method, a generic method for obtaining technical conditions $\T$, stability theorems such as  ($*$) and optimality intervals for loss-based clustering, from tractable convex relaxations. We will illustrate the working of this method for the K-means cost function (Sections \ref{sec:kmeans} and \ref{sec:exp-kmeans}), for graph partitioning by Normalized Cuts (Sections \ref{sec:ncut} and \ref{sec:experiments-ncut}), to any of a large class of losses (Section \ref{sec:other-btr}). Section \ref{sec:asymptotics} defines stability in the population setting. The relationship with previous work is presented in Section \ref{sec:related}, and the discussion in Section \ref{sec:discussion} concludes the paper.

\section{Preliminaries and definitions}
\label{sec:defs}
\paragraph{Representing clusterings as matrices}
Let $\dataset=\{x_{1},\ldots x_n\}$ be the data to be clustered. We
make no assumptions about the distribution of these data for now. A {\em
  clustering} $\clust=\{C_1,\ldots\, C_K\}$ of the data set $\dataset$ is a partition of the indices
$\{1,2,\ldots n\}\stackrel{def}{=} [n]$ into $K$ non-empty mutually disjoint subsets
$C_1,\ldots C_K$, called {\em clusters}. We denote by
$\clustal$ the space of all clusterings with $K$ clusters.

Let $n_k=|C_k|$ for $k=1,\ldots K$, with $\sum_{k=1}^Kn_k=n$; further, let $\pmin=\min_{k\in[K]}n_k/n,\,\pmax=\max_{k\in[K]}n_k/n$ represent the minimum, respectively maximum relative cluster sizes. 

A clustering $\clust$ can be represented by an $n\times n$ {\em clustering matrix} $X$ 
defined as
\beq\label{eq:X}
X\;=\;[X_{ij}]_{i,j=1}^n,
\quad
X_{ij}\;=\;\left\{\begin{array}{ll} 1/{n_k} & \text{if } i,j\in C_k 
         \,\text{for some $k\in [K]$}\\
                                   0 & \text{otherwise.}\\
           \end{array}\right.
\eeq
The following proposition lists the properties of the matrix $X$. The proof (which is straightforward) can be found in the Appendix, along with all other proofs.
\begin{prop}\label{prop:X}
For any clustering $\clust$ of $n$ data points, the matrix $X$ defined by \eqref{eq:X} has elements $X_{ij}$ in $[0,1]$; $\trace X=K$, %where $K=|\clust|$;  
$X\bfone=\bfone$, where $\bfone=[\,1\,\ldots\, 1\,]^T\in\rrr^n$, and $\|X\|^2_F\stackrel{def}{=}\trace X^TX=K$. Moreover, $X\succeq 0$, i.e. $X$ is a 
positive semidefinite matrix.% (i.e. $X$ symmetric and $x^TXx\geq 0$ for all vectors $x$).
\end{prop}
To distinguish a clustering matrix $X$ from other $n\times n$ symmetric matrices satisfying Proposition \ref{prop:X}, we sometimes denote the former by $X(\clust)$.\mmp{, thus emphasizing that it corresponds to a clustering $\clust$.} 
\paragraph{Measuring the distance between two clusterings} 
The {\em earth mover's distance} (also called the {\em misclassification error distance}) between two clusterings $\clust,\clust'$ over the same set of $n$ points is
\beq\label{eq:ME}
\distance(\clust,\clust')\;=\;1-\frac{1}{n}\max_{\pi\in{\mathbb S}_K}\sum_{k=1}^K \left(\sum_{i\in C_k\cap C'_{\pi(k)}} \!\!\!\!1\right),
\eeq
where $\pi$ ranges over the set of all permutations of $K$ elements ${\mathbb S}_K$, and $\pi(k)$ indexes a cluster in $\clust'$. If the data points have  weights $w_i,\,i\in [n]$, the {\em weighted earth mover's distance} is 
\beq\label{eq:MEw}
\distance_{w}(\clust,\clust')\;=\;1-\frac{1}{\sum_{i=1}^nw_i}\max_{\pi\in{\mathbb S}_K} \sum_{k=1}^K\left(\sum_{i\in C_k\cap C'_{\pi(k)}}\!\!\!\!w_i\right)\,.
\eeq
In the above, $\sum_{i\in C_k\cap C'_{k'}}\!\!w_i$ represents the total
weight of the set of points assigned to cluster $k$ by $\clust$ and to
cluster $k'$ by $\clust'$\footnote{These definitions can be generalized to
clusterings with different numbers of clusters, but here we will not
be concerned with them.}.

\paragraph{Losses and convex relaxations}
A {\em loss function} $\loss(\dataset,\clust)$ (such as the
K-means or K-medians loss) specifies what kind of clusters the user is interested in via the optimization problem below.
\beq \label{eq:clustering}
\mbox{ Clustering problem:} \
\quad L\opt\;=\;\min_{\clust\in \clustal}\loss(\dataset,\clust),\quad\text{with \mf{global} solution } \clust\opt
\eeq
%\end{problem}
\mf{in above problem $\clust'$ should be $\clust$?} As most loss functions require a number
of clusters $K$ as input, we assume that $K$ is fixed and given. In
Section \ref{sec:discussion} we return to the issue of choosing $K$.
The majority of interesting loss functions result in combinatorial optimization problems \eqref{eq:clustering} known to be hard in the worst case. 
A {\em convex relaxation} of the problem \eqref{eq:clustering} is an optimization problem defined as follows.
Let $\X$ be a convex set in a Euclidean space, such that $\X\supset\{X(\clust),\,\clust\in\clustal\}$. Extend $\loss(\dataset,\clust)$ to $\loss(\dataset,X)$, convex in $X$ for all $X\in\X$. Then,
\beq\label{eq:convexrel}
L^*\;=\;\min_{X\in \X}\loss(\dataset,X),\quad \text{with solution }X^*
\eeq
is a convex relaxtion for the clustering problem~\eqref{eq:clustering}.
In the above, the representation $X(\clust)$ can be the one defined in \eqref{eq:X}, or a different injective mapping of $\clustal$ into a Euclidean space. Because $\X\supset \clustal$, we have $L^*\leq L\opt$ and $X^*$ is generally not a clustering matrix.

Convex relaxations for clustering have received considerable interest.
For graph partitioning problems, \cite{Xing:CSD-03-1265} introduced
two relaxations based on Semi-Definite Programs (SDP).  Correlation
clustering, a graph clustering problem appearing in image analysis,
has been given an SDP relaxation in \cite{DBLP:conf/soda/Swamy04} and
\cite{DBLP:conf/icalp/AhmadianS16}. For community detection under the
Stochastic Block Model \citep{holland1983stochastic} several SDP
relaxations have been recently introduced by \cite{chen2014statistical},
\cite{vinayak2014graph} and \cite{JHDF:16} as well as Sum-of-Squares
relaxations for finding hidden cliques, in \cite{2015arXiv150206590D}. For
centroid based clustering, we have Linear Program (LP) based relaxations for K-medians
by \cite{charikar:99} and K-means \cite{awasthiCKS:15} and more recent,
tighter relaxations via SDP in \cite{Aswathi+Ward:14}. The SDP
relaxations of \cite{Aswathi+Ward:14,Iguchi:15} for K-means have
guarantees under the specific generative model called the Stochastic
Ball Model \citep{Iguchi:15b}. Relaxations exist also for
exemplar-based clustering \citep{DBLP:conf/nips/ZhuXLY14}.
For hierarchical clustering in the cost-based paradigm introduced by
\cite{Dasgupta:16}, we have LP relaxations introduced by
\cite{royPokutta:hierarchical16,charikar:99,charikar:16}.\comment{We also
mention the related area of {\em convex clustering}
\citep{DBLP:conf/nips/BachH07} where loss functions are designed to be
convex. These can often be seen as relaxations to standard K-center or
exemplar based losses.}

\section{The Sublevel Set method: proving stability via convex relaxations}
\label{sec:generic}
%\subsection{Sublevel set problems and the generic Sublevel Set method}
\label{sec:bound-pb}
Now we show how to use an existing relaxation to obtain guarantees of the form \probstar~ for clustering. Given a \loss, its clustering problem \eqref{eq:clustering}, and a convex relaxation \eqref{eq:convexrel} for it we proceed as follows:
Step 1: We use the convex relaxation to find a set of good clusterings that contains a given $\clust$ \mf{not clear which $\clust$}. This set is $\X\llll=\{X\in\X,\,\loss(\dataset,X)\leq l\}$ \mf{$\clust$ here in Loss?}, the {\em sublevel set} of \loss, at the value $l=\loss(\dataset,\clust)$.  This set is convex when \loss~is convex in $X$.
Step 2: We show that if $\X\llll$ is sufficiently small, then all
  clusterings in it are contained in the $\distance$ $\epsilon$-ball
  $\{\clust',\;\distance(\clust,\clust')\leq \epsi\}$. This ball is
  an optimality interval for $\dataset, \loss$ and $K$. 

In more detail, consider a dataset $\dataset$, with a clustering
$\clust\in\clustal$. Assume for the given \loss~a convex relaxation
exists, with feasible set $\X$, and let $X(\clust)$ be the image of
$\clust$ in $\X$. We modify the relaxed optimization problem
\eqref{eq:convexrel} to define optimization problems such as the one below, which we call {\em Sublevel Set (SS)} problems.
 \beq \label{eq:bpb1}
\text{{SS} Problem}\quad \epsi'\;=\;\max_{X'\in
  \X}\|X(\clust)-X'\|,\quad\text{s.t. } \loss(\dataset,X')\leq
\loss(\dataset,\clust).
\eeq
The norm $||\,||$ can be chosen conveniently and in all cases presented here it will be
the Frobenius norm $||\,||_F$ defined in Proposition \ref{prop:X}. The
feasible set for \eqref{eq:bpb1} is
$\X\llle{\loss(\dataset,\clust)}$, a convex set. The convexity and
tractability of this SS problem depends on its objective, and we
will show that the mapping $X(\clust)$ in \eqref{eq:X}, along with the
Frobenius norm, always leads to tractable SS problems; we will present
other examples of such mappings in Section \ref{sec:other-btr}.  When
the SS problem is tractable, then, by solving it we obtain that
$||X(\clust')-X(\clust)||\leq \epsi'$ for all clusterings $\clust'$
with $\loss(\dataset,\clust')\leq
\loss(\dataset,X(\clust))$. 
\begin{prop}[An alternative SS problem]\label{prop:epsi2} Let SS2 %\beq \label{eq:bpb2}
 be the problem $\epsi'_2\;=\;\max_{X'\in \X}\|X^*-X'\|,\quad\text{s.t. } \loss(\dataset,X')\leq \loss(\dataset,\clust)$, and $\epsi'$ be defined as in \eqref{eq:bpb1}. From SS2, we obtain that (1) for each clustering $\clust'$, $\loss(\dataset,\clust')\leq\loss(\dataset,\clust)$ implies $||X(\clust')-X(\clust)||\leq 2\epsi'_2$, and (2), that  $\epsi'\leq 2\epsi_2'$.
\end{prop}
Hence, the bound $\epsi_2'$ is never tighter than $\epsi'$; moreover, $\epsi'_2$ requires  extra computation to obtain $X^*$ from \eqref{eq:convexrel}. Therefore, from now on we focus solely on the SS problem.
 
The optimal value $\epsi'$ of SS defines a ball centered at $\clust$ that contains all the good clusterings. Throughout this section, ``good'' is defined as ``at least as good as $\clust$'' (w.r.t. \loss), but any other level $l$ can be considered, even levels $l<\loss(\dataset,\clust)$. The radii of the corresponding sublevel sets tell us how clusterable the data is. 

The value $\epsi'=||X(\clust')-X(\clust)||$ could be considered a distance between partitions, but this distance is less intutive, and has the added disadvantage that it depends on the mapping $X$ used. In Step 2, we transform the bound $\epsi'$ into a bound on the earth mover's distance $\distance$, using the following result.
\begin{thm}[\cite{M:equivalence-ML10}, Theorem 9]\label{thm:equivalence-ML10}
For two clusterings $\clust,\clust'$ with the same number of clusters $K$, denote $\clust=\{C_1,\ldots \C_K\}$, and $\pmin=\frac{1}{n}\min_{[K]}|C_k|$, $\pmax=\frac{1}{n}\max_{[K]}|C_k|$. Then, for any $\epsi\leq \pmin$, if $\frac{1}{2}\|X(\clust)-X(\clust')\|^2_F\leq\frac{\epsi}{\pmax}$, then $\distance(\clust,\clust')\leq\epsi$. 
\end{thm}
\mmp{the bound is not tight, but the proof of thm 9 shows what is needed to make it tight, and we should do it for the next paper}

The remainder of this section specializes the SS method to some popular clustering loss functions.

\subsection{Optimality intervals for the K-means loss}
\label{sec:kmeans}
In K-means clustering, the data are $\dataset = \{x_1,x_2,\cdots,x_n\}\in\rrr^d$. The objective is to minimize the {\em squared error loss}, also known as the {\em K-means loss} 
\beq\label{eq:loss0}
\lossk(\dataset,\clust)\;=\;\frac{1}{n}\sum_{k=1}^K\sum_{i\in C_k}||x_i-\mu_k||^2,
\quad \text{with }
\mu_k\;=\;\frac{1}{n_k}\sum_{i\in C_k}x_i,\quad \text{for $k\in[K]$}.
\eeq
Define the {\em squared distances matrix} $D$ by
\beq\label{eq:D}
D\;=\;[D_{ij}]_{i,j\in[n]},\quad \text{ } D_{ij}\;=\;||x_i-x_j||^2
\eeq
where  $||x||$ denotes the Euclidean norm of $x$. Furthermore, let $\langle A,B\rangle\stackrel{def}{=}\trace(A^TB)$ denote the Frobenius scalar product and recall that $||A||_F=\langle A,A \rangle^{1/2}$.
\comment{ If one substitutes the expressions of the centers $\mu_{1:K}$ into \lossk, one can express it} It can be shown that \lossk~is a function of the matrices $X$ and $D$.
\beq\label{eq:loss}
\lossk(\dataset,\clust)\;\equiv\;\lossk(D,X(\clust))\;=\;\frac{1}{2n}\langle D,X(\clust)\rangle.
\eeq
This formulation inspired \cite{peng:07} to propose the following convex relaxation of the K-means problem
\comment{\beq \label{eq:kmeans-pb}
\min_{\clust\in\clustal} \langle D,X(\clust)\rangle
\eeq
}\comment{several tractable relaxations for the K-means problem have been developed; \cite{ding:04,dhillon:04} introduced a spectral relaxation, \cite{Aswathi+Ward:14}\mmp{others before} introduced two convex relaxations, one resulting in a Linear Program (LP), the other in}
\beq
\label{eq:kmeans-sdp}
\min_{X\in\X}\langle D,X\rangle
\eeq
where $\X=\{X\in\rrr^{n\times n},\,\trace X=K, \,
X\bfone=\bfone,\,X_{ij}\geq0,\,\text{for }i,j\in[n],\,X\,\succeq\,0\}$
is the set of matrices satisfying the conditions in Proposition
\ref{prop:X}. In \citep{peng:07} it was shown that problem
\eqref{eq:kmeans-sdp} can be cast as a {\em Semidefinite Program (SDP)}.
\comment{moved to experiments: In general, $X^*$ the optimal solution
  of \eqref{eq:kmeans-sdp} is not a clustering
  matrix. \cite{Aswathi+Ward:14} showed that when data are sampled
  from well-separated discs, $X^*$ is a clustering matrix
  corresponding to the optimal clustering $\clust^*$ of the data (and
  $\clust^*$ assigns the points in each disk to a different cluster).}

%\subsection{The Sublevel Set problem for K-means}
%\label{sec:kmeans-bound}
We use the relaxation \eqref{eq:kmeans-sdp} to obtain OI for K-means.
We shall assume that a data set $\dataset$ is given, and that the user has already found a clustering $\clust$ of this data set (by e.g. running the K-means algorithm).\comment{ The user would like to know if: (a) is $\clust$ optimal (in other words, is it the globally optimal solution to the non-convex problem \eqref{eq:kmeans-pb})? and (b) could there be other clusterings of the data $\dataset$ that are very different from $\clust$ but are similar or better w.r.t to $\loss$? As Figure \ref{fig:kmeans-intuition} shows, the two questions are both important, if the goal of the clustering is to capture the structure of the data (i.e. the clustering) instead of minimizing the clustering $\loss$. }
\comment{There can be optimal clusterings that are not stable; we cannot prove that clust is optimal, but we can prove that is close to optimal by stabilty}
 The SDP below corresponds to the SS  problem from Section \ref{sec:generic}. The main difference from equation \eqref{eq:bpb1} is that we used the identity $\|X(\clust)-X'\|^2=2K-2\langle X(\clust),X'\rangle$ to obtain a convex minimization objective instead of a norm maximization.
\beq\label{eq:kmeans-probe}
\probe\quad \kappa(\clust)\;=\;\min_{X'\in\X}\langle X(\clust),X'\rangle\quad\text{s.t.} \langle D,X'\rangle\,\leq\, \langle D,X(\clust)\rangle
\eeq
\comment{In the above, $X(\clust)$ is the clustering matrix of the known clustering, $\X$ is defined above\comment{in \eqref{eq:kmeans-sdp}} and $D$ is the squared distance matrix of $\dataset$.} 
\comment{REMOVE THIS? Let us now examine what the optimal solution $X'$ and optimal value $\kappa(\clust)$ mean.  In comparison with \eqref{eq:kmeans-sdp}, $\probe$ adds an inequality constraint, thus restricting the feasible set of \eqref{eq:kmeans-sdp} to matrices $X'$ that have $\lossk$ no larger than the loss of $\clust$.  Both $X^*$ and $X(\clust)$ are feasible for $\probe$, but clusterings with higher $\loss$ than $\clust$ are not. 
Hence,  $\probe$  finds among the feasible matrices $X'$ which have low loss, the one that is furthest away from $X(\clust)$. As Theorem \ref{prop:mebound} below will show, the optimum value $\kappa$ in $\probe$ determines this distance, measured in Frobenius norm. Consequently, if for the given data and $\clust$ the value $K-\kappa$ is small enough, it implies that no good clusterings of the data can differ more than $K-\kappa $ from $X(\clust)$. \comment{The Frobenius metric between clustering matrices is a valid distance between partitions (see e.g. \cite{back,arabie}), however it is less intuitive than the {\em earth mover distance} (or {\em misclassification error} distance) $d$ (defined below/above).}} Our main result below states that when the value $\kappa(\clust)$ is near $K$, it controls the maximum deviation from $\clust$ of any other good clustering. 
\begin{thm}\label{prop:mebound} Let $\dataset$ be represented by its squared distance matrix $D$, let $\clust$ be a clustering of $\dataset$, with $K,\pmin,\pmax$ as in Section \ref{sec:defs}, and let $\kappa(\clust)$ be the optimal value of problem $\probe$. Then, if $\epsi=(K-\kappa(\clust))\pmax\leq \pmin$, any clustering $\clust'$ with $\loss(\clust')\leq \loss(\clust)$ is at distance $\distance(\clust,\clust')\leq \epsi$.
\end{thm}
When $\epsi$ defined by Theorem \ref{prop:mebound} is smaller than the relative size of the smallest cluster, then  $\clust$, even though not necessarily optimal, is a representative of a small set that contains the optimal clustering $\clust\opt$ as well as all the other clusterings that are as good as $\clust$. Sometimes, when $\epsi<\frac{1}{n}$, as in Figure \ref{fig:example-kmeans}, Theorem \ref{prop:mebound} also implies that $\clust=\clust\opt$. 
With  $\clust$ and $D$ known, a user can solve this SDP in practice and obtain an OI defined by $\epsi$. We summarize this procedure below.
\begin{center}
\fbox{
\begin{minipage}{\textwidth}
\benum
\item[]{\bf Input} Data set with $D\in\rrr^{n\times n}$ defined as in \eqref{eq:D}, clustering $\clust$ with $K$ clusters, $\pmin,\pmax$, and clustering matrix $X(\clust)$. 
\item Solve problem $\probe$ numerically (by e.g. calling a SDP solver); let $\kappa$ be the optimal value obtained. 
\item Set $\epsilon=(K-\kappa)\pmax$.
\item {\bf If} $\epsilon\leq \pmin$ {\bf then}
   \bit
   \item[] Theorem \ref{prop:mebound} holds: $\epsi$ gives an OI for $\clust$.
   \eit
\item[]{\bf else} no guarantees for $\clust$ by this method. 
\eenum
\end{minipage}
}% end fbox
\end{center}
%
%\paragraph{Discussion}
The above method exemplifies the goals set forth in the Introduction; it depends only on observed and computable quantities, and does not relie on assumptions about the data generating process. These bounds exist only when the data is clusterable. Currently we cannot show that all the clusterable cases can be given guarantees; this depends on the tightness of the relaxation, as well as on the tightness of Step 2 of the SS method. 

%\label{sec:kmeans-lp}
\hanyuz{Need to mention this?}  The SDP relaxation \eqref{eq:kmeans-sdp} is not the only way to obtain a SS problem for $\lossk$, and we further illustrate the versatility of the SS method by constructing a second SS problem for this clustering loss.
In \cite{Aswathi+Ward:14} the following relaxation
to the K-means problem is presented.
\beq \label{eq:kmeans-lp}
\min_{X\in\X\lp} \langle D,X\rangle.  \eeq
In the above, the mapping
$\clust\rightarrow X(\clust)$ is the same as in \eqref{eq:X}; the
convex set $\X\lp$ is
%\beq
$\X\lp=\{\trace X=K,\,X\bfone=\bfone,\,X_{ij}\leq X_{ii}\,\text{for all } i,j\in [n],\,X_{ij}\in[0,1]\,\text{for all } i,j\in [n]\}.$
%\eeq
%
 Relaxation \eqref{eq:kmeans-lp} can be cast as a Linear Program, making it more attractive from the computational point of view. It is straightforward to state the Sublevel Set problem SS corresponding  to \eqref{eq:kmeans-lp}, which is also an LP.
\beq \label{eq:kmeans-lp-bound}
\kappa\lp(\clust)\;=\;\min_{X'\in\X\lp}\langle X(\clust),X'\rangle,\quad \text{s.t. }
\langle D,X'\rangle\;\leq\; \langle D,X\rangle.
\eeq
Since $\kappa\lp(\clust)$ bounds the same quantity $\langle X(\clust),X'\rangle$, Theorem \ref{prop:mebound} applies. When multiple OI can be obtained, the tightest
one bounds the distance $\distance(\clust, \clust')$.\comment{ Unfortunately,
  in this case $\kappa\lp\leq \kappa$ always, due to the following
  result.
\begin{prop} Let $\X$ be defined as in \eqref{eq:kmeans-sdp}. Then $\X\lp\supset\X$.
\end{prop} NOT TRUE, I think. true if $X_{ij}^2>X_{ii}$, which never holds.
}
In \cite{Aswathi+Ward:14} it is shown that the SDP relaxation is strictly tighter than the LP relaxation, for data generated from separated balls. This suggests that the OI from the LP will not be as tight as the SDP OI.

\subsection{Optimality intervals for Normalized Cut graph partitioning}
\label{sec:ncut}
Now we exemplify the Sublevel Set method with graph partitioning under the {\em Normalized Cut} loss. The data consists of a weighted graph $G=([n],W)$, where $W=[W_{ij}]_{i,j\in[n]}$ is a symmetric matrix with non-negative entries; $W_{ij}>0$ means that there exists an edge between nodes $i$ and $j$ whose weight is $W_{ij}$. The {\em degree} of node $i$ is defined as $w_i=\sum_{j=1}^nW_{ij}$. We denote by $w\in\rrr^n$ the vector of all degrees, and by $\pmin (\pmax)$ $\min (\max)_{[K]}(\sum_{i\in C_k}w_i)/(\sum_{i\in[n]}w_i)$. The Normalized Cut loss can be defined following \cite{MShi:aistats01,ding:04} as 
\beq \label{eq:NCut}
\loss\ncut(W,\clust)=\sum_{k\in[K]}\left(\sum_{i\in C_k,j\not\in C_k}W_{ij}\right)/\left(\sum_{i\in C_k}w_i\right).
\eeq
Minimizing \loss\ncut~ is provably NP-hard \cite{ShiMalik_ncut_pami:00} for any $K\geq 2$. 
\cite{Xing:CSD-03-1265} introduced the convex SDP relaxation
\beq \label{eq:ncut-sdp}
\min_{X\in\X\ncut} \langle L,X \rangle,
\eeq
with $L=I-\diag(w)^{-1/2}W\diag(w)^{-1/2}$, over the space
\beq
\X\ncut\;=\;
\{X \succeq 0,\,\trace X=K,\, X\!\diag(w)^{\!\!-1/2}\!=\!\diag(w)^{\!\!-1/2},\,I-X\succeq 0,X\geq 0\}
\eeq
In this relaxation, a clustering $\clust$ is mapped to
\beq\label{eq:Xncut}
X_{ij}(\clust)\;=\;\frac{\sqrt{w_iw_j}}{\sum_{i'\in\clust_k}w_{i'}}\;\text{if } i,j\in C_k,\;\text{and 0 otherwise, for all $i,j\in[n]$.} 
\eeq
The SS problem based on this relaxation is
\beq \label{eq:ncut-probe}
\probncut \quad\kappa(\clust)\;=\;\min_{X'\in \X_{\ncut}} \langle X(\clust),X'\rangle,\quad
\text{s.t. } \langle L,X'\rangle\,\leq\,\langle L,X(\clust)\rangle.
\eeq
We have the following result
\begin{thm}\label{prop:ncut} Let $G=([n],W)$ be defined as above, $\clust$ be a clustering of $[n]$, and $\kappa(\clust)$ be the optimal value of problem \eqref{eq:ncut-probe}. Let $\epsi=(K-\kappa(\clust))\pmax$ Then, if $\epsi\leq \pmin$, any clustering $\clust'$ with $\loss_{\ncut}(\clust')\leq \loss_{\ncut}(\clust)$ is at distance $\distance_{w_{1:n}}(\clust,\clust')\leq \epsi$. 
\end{thm}
The proof is similar to the proof of Theorem \ref{prop:mebound} and is sketched in the Appendix.

\subsection{For what other clustering paradigms can we obtain optimality intervals?}
\label{sec:other-btr}
%Furthermore,  we show that the Sublevel Set method of Section \ref{sec:generic} can be applied to a multitude of clustering paradigms with very little extra work. 
Define the following injective mappings of $\clustal$ into sets of
matrices. The $X$ mapping is given in \eqref{eq:X}. The mapping
$\tilX:\clustal\to\rrr^{n\times n}$ is given by $\tilX_{ij}(\clust)=1$ if $i,j\in C_k$
for some $K$ and 0 otherwise. The mapping $Z:\clustal\to\rrr^{n\times K}$ is
given by $Z_{ij}(\clust)=1/\sqrt{n_k}$ if $i\in C_k$ for $k\in [K]$ and 0
otherwise. Define the spaces $\tilde{\X},\mathcal{Z}$ by
\beq \label{eq:tildeXspace}
\tilde{\X}\;=\;\{\tilX\in\rrr^{n\times n},\,||\tilX||^2_F\leq (n-K+1)^2+K-1,\,\tilX\succeq 0,\,\tilX_{ij}\in[0,1]\,\text{for }i,j\in[n]\}
\eeq
and
\beq \label{eq:Zspace}
\mathcal{Z}\;=\;\{Z\in\rrr^{n\times K}, \,\text{orthogonal}\}
\eeq
It is easy to verify that $\tilX(\clust)\in \tilde{\X}$, respectively that $Z(\clust)\in {\mathcal Z}$ for any $\clust\in\clustal$.
\begin{thm}\label{prop:generic-XZ}
Let \loss~be a clustering loss function that has a convex
relaxation. If in this relaxation a clustering $\clust$ is mapped to one of the
matrices $X(\clust),\tilX(\clust),Z(\clust)$ above, then the following statements hold.
\bit
\item[(1)] The SS problem 
$\kappa=\min_{X'\in \X\llll}\langle X(\clust),X'\rangle$
(and similarly for $\tilX(\clust),Z(\clust)$) has convex sublevel sets $\X\llll,\tilde{{\mathcal X}}\llll,{\mathcal Z}\llll$ for any $l$.  
\item[(2)]%  From the optimal value $\kappa$ an OI $\epsi$ can be obtained as follows.
For the $X$ mapping, let $\epsi=(K-\kappa)\pmax$; for the
$\tilX$ mapping, let
$\epsi=\frac{\sum_{k\in[K]}n_k^2+(n-K+1)^2+(K-1)-2\kappa}{2\pmin}$;
for the $Z$ mapping, let $\epsi=(K-\kappa^2/2)\pmax$. In all three
cases, $\epsi$ is an OI whenever $\epsi\leq \pmin$.
\eit
\end{thm}
Of the previously mentioned relaxations the  $X$ mapping is used by \cite{peng:07,Iguchi:15} for K-means in a SDP relaxations, by \cite{royPokutta:hierarchical16,charikar:16} for cost-based hierarchical clustering in an LP relaxation, and by \cite{DBLP:conf/soda/Swamy04} for correlation clustering. The $\tilX$ mapping is used by \cite{chen2014statistical,JHDF:16} for the Stochastic Block Model \citep{holland1983stochastic}, respectively by \cite{vinayak2014graph} for the Degree-Corrected Stochastic Block Model \cite{karrerN:dcsbm11}. The $Z$ mapping is used by the spectral relaxation of K-means by \cite{ding:04}. Finally, note that the relaxations in \cite{heinS:nips11,rangapuranPramodHein:nips14} are not covered by Theorem \ref{prop:generic-XZ}.

Theorem \ref{prop:generic-XZ} can also be extended to cover weighted representations such as those used for graph partitioning in \cite{MShortreedXu:aistats05}. Theorem \ref{prop:generic-XZ} shows, somewhat counterintuitively, that getting bounds for a clustering paradigm {\em does not depend directly on the  $\loss$},  but on the space of the convex relaxation. Moreover, somebody who already uses one of the above cited relaxations to cluster data would have very little additional coding work to do to also obtain optimality intervals.

\section{Population stability for the K-means loss}
\label{sec:asymptotics}

\newcommand{\distancep}{d_{\Pp}^{EM}}
\newcommand{\distancehat}{d_{\widehat{Pp}}^{EM}}
It is natural to ask if stability of a clustering $\clust$ on a sample $\dataset$ can allow us to infer something about the distribution that generated the sample. This section shows that this is indeed possible, with only generic Glivenko-Cantelli type assumptions on $\lossk$. 

\hanyuz{As I re-read and try to re-write this section, this seems to be more and more trivial to me :(.  }

Throughout this section, we assume the following conditions to be true.
\begin{assumption}
The data $\dataset=\{x_1,\cdots,x_n\}$ is sampled i.i.d. from $\Pp$, a distribution supported on a subset of $\rrr^d$. $\Pp$ is absolutely continuous with respect to the Lebesgue measure on $\rrr^d$.
\end{assumption}

We start by expanding the relevant definitions to the population case.
For K-means, a clustering $\clust$ is a partition of $\rrr^d$ induced by $K$ {\em Voronoi centers} $\mu_1,\cdots,\mu_K \in \rrr^d$; every $x\in\rrr^d$ is assigned a label with the closest center.\footnote{$\clust$ is defined only up to a zero measure set,  but we can ignore such distinctions since $\lossk$ and $\distance_\Pp$ as defined in this section\comment{\eqref{eq:loss0-population}, respectively \eqref{eq:distance-P}} are invariant to them.}  We then identify each clustering $\clust$ with the set of its Voronoi centers. This ensures that $\lossk(\Pp,\clust)$ is well defined
\begin{equation} \label{eq:loss0-population}
\lossk(\Pp,\clust) = \int_{\rrr^d} \min_{k\in[K]}\norm{x-\mu_k}^2 \Pp(\mathrm{d}x)
\end{equation}
\mmp{The mapping $\mu\to\clust\in \clustal(\rrr^d)\to\clust \in \clustal(\dataset)$ is well defined/unique. The reverse is not. In particular, for $K<d+1$, I think, $\clust$ does not uniquely determine the centers. But this does not matter, because for the optimum, they are always uniquely defined. TO KEEP THIS HERE AS A REMINDER}
Denote by $\clustal(\rrr^d)$,  $\clustal(\dataset)$ the set of all clusterings of $\rrr^d$, respectively of a fixed $\dataset$, defined by  $K$ {\em distinct} Voronoi centers.\footnote{ The reader will note that $\clustal(\dataset)$ is only a subset of $\clustal$ of Section \ref{sec:defs}. We need this restriction to ensure that $\clustal(\rrr^d)$ has finite VC-dimension. Moreover, the mapping from Voronoi centers to partitions of $\rrr^d$ is not injective; however, this does not affect the results in this paper.} With a slight abuse of notation, we will use $\clust=\{C_1,\ldots C_K\}$ for clusterings in either set.
\mmp{In the finite sample case, $\clustal(\dataset)$ is redefined to include only clusterings defined by $K$ distinct Voronoi centers. Hence, any such set of centers will define a Voronoi partition $\clust\in\clustal(\rrr^n)$ and a partion of $\dataset$ (assumed fixed) into sets of points in $\clustal(\dataset)$. }
 Minimizing $\lossk(\dataset,\clust)$, as defined in \eqref{eq:loss0-population}, when we view $\dataset$ as a population with \emph{finite} support leads to the previous definition of K-means loss in \eqref{eq:loss0}. Note however that, for an arbitrary $\clust\in \clustal(\rrr^d)$ or $\clustal(\dataset)$, the Voronoi centers do not coincide with the means of the clusters, unless $\clust$ is a fixed point of the K-means algorithm.

The following assumption is made about $\Pp$.
\begin{assumption}[Uniform Convergence of $\lossk$] \label{ass:uniform} There exists a function $\Psi$ such that, for any $n$ sufficiently large and any $\delta\in(0,1]$,  with probability $1-\delta$ over resampling the size $n$ sample $\dataset$ from $P$,
\begin{equation}\label{eq:distance-P}
\sup_{\clust\in\clustal(\dataset)}
|\lossk(\Pp;\clust)-\lossk(\dataset,\clust)| \leq \Psi(n,\delta)\,,
\end{equation}
\end{assumption}
In the above, the supremum is taken over all sets of distinct Voronoi centers, which allows an identification of a $\clust\in\clustal(\rrr^d)$ from $\clust\in\clustal(\dataset)$. Intuitively, equation \eqref{eq:distance-P} bounds the difference between $\lossk(\dataset,\clust)$ and the $\lossk$ of any clustering of $\rrr^n$ that is consistent with $\clust$.
Assumption \ref{ass:uniform} holds, for instance, when $\Pp$ has compact support \citep{k-dim-2010-Maurer} or finite higher order moments \citep{Telgarsky-bound}. We now view $\Psi(n,\delta)$ as a known function of $n,\delta$.

The earth mover's distance can be directly generalized to $\clustal(\rrr^d)\times \clustal(\rrr^d)$ as the limit when $n\to \infty$ of $\distance_w(\clust,\clust')$. We will denote with $\distance_{\Pp}(\clust,\clust')$ the distance of two clusterings $\clust,\clust'\in\clustal(\rrr^d)$, and $\distance(\clust,\clust')$ as before for distances of two clusterings in $\clustal(\dataset)$.

We now expand the definition of $\epsi$-stability to include a parameter $\Delta$.
%\begin{definition}[$(\Delta, \epsi)$ stable clustering]
A clustering $\clust\in\clustal(\rrr^d)$ is called $(\Delta,\epsi)$ stable if any clustering $\clust'\in\clustal(\rrr^d)$ with $\loss(\Pp,\clust')\leq \loss(\Pp,\clust)+\Delta$ is at distance $\distance_{\Pp}(\clust,\clust')\leq \epsi$.
%\end{definition}
A similar definition holds for $(\Delta,\epsi)$-stable clusterings in $\clustal(\dataset)$.  If $\clust$ is not $(\Delta,\epsi)$ stable then it is called $(\Delta,\epsi)$ unstable. Note that $(\Delta,\epsi)$-stability (or instability) for any clustering $\clust$ implies (weaker) stability (or instability) statements for any other clustering $\clust'$ in the $l=\loss(\clust)+\Delta$ sublevel set. For example, if $\clust$ is $(\Delta,\epsi)$ stable, and $\loss(\clust')=\loss(\clust)+\Delta'$, with $\Delta'<\Delta$, then $\clust'$ is $(\Delta-\Delta',2\epsi)$-stable.
\hanyuz{Note that for a continuous loss, as long as the global optimum is unique, there exists a pair of $(\Delta,\epsi)$ such that the population is $(\Delta,\epsi)$ stable. This was a property that holds for all absolute continuous population. ? not sure how to expand this point.}
Furthermore, the following family of SS problems parametrized by the excess loss $\Delta$ can be used to verify $(\Delta,\epsi)$ stability on a sample $\dataset$.
\begin{equation} \label{eq:kmeans-sdp-delta}
       \probe(\Delta) \quad \kappa(\Delta) = \min_{X'\in\X}\langle X(\C),X' \rangle \quad s.t. \langle D,X'\rangle \leq \langle D,X(\C) \rangle  + \Delta,
    \end{equation}
In the above,  $\X$ is defined by \eqref{eq:X} as in Section \ref{sec:kmeans}. Obviously, \probe(0) is identical to the K-means SS problem \eqref{eq:kmeans-sdp}. Moreover, all the results in Section \ref{sec:kmeans} can be generalized for  $\probe(\Delta)$ as well. In other words, for any $\dataset, \clust$ and $\Delta$, one can obtain an optimality interval from $\probe(\Delta)$ whenever the resulting $\epsi=(K-\kappa(\Delta))\pmax$ is no larger than $\pmin$.

The following theorem shows how stability guarantees obtained from \probe$(\Delta)$ in a sample $\dataset$ can support stability inferences in the distribution $\Pp$.
 \begin{thm}
 Suppose $\Pp$ satisfies Assumptions 1 and 2, and let $\delta\in(0,1]$. If any optimal clustering $\clust^{opt}$ on $\Pp$ is $(\Delta,\epsi)$ unstable for some $\Delta > 0$, then with probability $1-\delta$ over samples $\dataset$, with $|\dataset|=n$, any optimal clustering  $\widehat{\clust}^{opt}$ of $\dataset$ is $(\Delta+2\Psi(n,{\delta}/{2}),\epsi/2-\sqrt{\log(4/\delta)/2n})$ unstable.
 \label{thm:instable}
 \end{thm}
This result opens the way for inferences on the $(\Delta,\epsi)$ clusterability of $\Pp$ that could be framed as a family of hypothesis tests parametrized by $\Delta$. Select $\delta\in (0,1]$ and $\Delta$ a tolerance of excess loss. Then consider null hypothesis
\begin{align*}
    H_0(\epsi): &\text{ Any optimal K-means clustering on $\Pp$ is $(\Delta,\epsi)$ instable. }
\end{align*}
Let $\kappa$ be the optimal value of solving $\probe(\Delta+2\Psi(n,\delta/2))$ on the sample $\dataset$. We reject the null hypothesis $H_0(\epsi)$ with $\epsi=2((K-\kappa)\pmax+\sqrt{\frac{\log(4/\delta)}{2n}})$ when $(K-\kappa)\pmax\leq \pmin$. \hanyuz{We need to check these two numbers: first is there a missing factor 2 as in previous propositions, second do we plus the VC bound when we want to reject the null hypothesis? i.e. I think we should reject when $\epsi \leq \pmin$}  Supposing $H_0(\epsi)$ is true, by Theorem \ref{thm:instable} the probability of type I error is at most $\delta$. Thus, one can interpret an OI from $\probe(\Delta+2\Psi(n,\delta/2))$ as sufficient for rejecting $(\Delta,\epsi)$ instability with a p-value at most $\delta$.\mmp{This should be in-clusterability,actually} Moreover, the inference above remain valid, albeit weaker, when instead of the optimal $\widehat{\clust}\opt$ only a sub-optimal clustering of the sample is known.  While this particular test would be overly conservative, and not necessarily practical, it serves to alert to the possibility of inferring stability in the population from finite sample stability.

Previously people have proposed and studied different paradigm of clustering stability \citep{ben-david:06} for model selection. However those notions fail to associate instability on sample with instability on population. The key difference between the assumptions we make and those of the previous papers, is the uniform bound $\Psi(n,\delta)$ for $\lossk$. Our result, on the other hand, shows that we could provide probability guarantee for the $(\Delta,\epsilon)$ stability in our framework for finite samples. More discussion about previous work on clustering stability can be found in \ref{sec:discussion}.

\mmp{this better in related work? in fact, i think we must add a discussion of all the work under this paradigm from ohad and ule. before this was not necessary because the paper was only on deterministc bounds}

\section{Related work}
\label{sec:related}
The first stability guarantees as defined by the generic Theorem $(*)$  were proposed in \cite{M:kmeans-distortion-icml06,MShortreedXu:aistats05}, with the OIs based on spectral bounds. This paper greatly expands the scope of \cite{M:kmeans-distortion-icml06,MShortreedXu:aistats05} to general tractable relaxations and to a much wider class of clustering problems, via the Sublevel Set method. In addtion, specific Sublevel Set problems and new OIs are obtained in the cases of the K-means and Normalized Cut losses, by using the Semidefinite Programming (SDP) relaxations.

\paragraph{Existing distribution free guarantees for clustering} All the previous explicit optimality intervals and associated bounds we are aware of are based on spectral relaxations: \cite{M:kmeans-distortion-icml06} gives a spectral OI for K-means and \cite{MShortreedXu:aistats05,MWan:nips15} give OI for graph partitioning under Normalized Cuts, respectively the Stochastic Block Model and extensions. The work of \cite{leeGharanTrevisan:14} relates the existence of good
$r$-way graph partitioning to a large
$K$-eigengap of the graph {\em normalized Laplacian}, where $r\geq K-3K\delta$. More precisely, if $\lambda_{K+K\delta}/\lambda_K>c(\log K)^2/\delta^9$ then this partition is ``better'' than
$\lambda_K/\delta^3\times c'$ (for $c,c'$ unspecified). While these
results are remarkable for their generality, they
require extremely large $\lambda_{K+K\delta}/\lambda_K$  to produce
non-trivial bounds, no matter what $c,c'$ are; moreover, because $\lambda_{K+K\delta}\leq 1$, they also require $\lambda_K\ll 1$. In \cite{pengSunZanetti:colt15}, an OI for spectral clustering is given, which depends on unspecified constants.

\paragraph{Algorithmic results under clusterability assumptions}
For finite mixtures, a series of results from the 2000's by
\cite{Dasgupta:00random}, \cite{achlioptas:05mixtures},
\cite{vempala:04mixtures}, \cite{dasgupta:07}, and a few more recent
ones by \cite{MBubeckLuxburg:esaim12},
and \cite{balakrishnanWY:arxiv14} established theoretical guarantees
for the approximate recovery of the original cluster membership by
tractable clustering algorithms. These papers are important because
for the first time, recovery is tied to the separation of the cluster centers,
and to the relative sizes and spreads of the clusters. Recovery
guarantees have been obtained also in {\em block-models} for network
data, such as the {\em Stochastic Block Model (SBM)} by
\cite{abbe2015community}, \cite{NIPS2016_6365}, the {\em
  Degree-Corrected SBM (DC-SBM)} by \cite{qinRohe:13} and the {\em Preference
  Frame Model (PFM)} by \cite{MWan:nips15}. 
Recovery results for graph clustering are given in e.g. \cite{kannan:00}. We
have already mentioned the recent \cite{Aswathi+Ward:14} and \cite{Iguchi:15}.

The SS methods are {\em complementary} to the work in this area. On
one hand, the cited works provide very strong evidence that if
$\dataset$ is clusterable, a good clustering $\clust$ is easy to
find. They corroborate a large body of empirical evidence, including
our own experiments in Section \ref{sec:exp-kmeans}. Since the SS
method is predicated on having found a good $\clust$, these works
suggest that the SS method will be applicable when the data is clusterable.\comment{ In future work we aim to close the loop
by providing end-to-end algorithms that both cluster data efficiently
and give BTRs for the resulting $\clust$.} In addition, the present paper
{\em grounds} the aforementioned area of research; by the SS method
one can hope to {\em prove the assumptions} that (some of) the algorithms
are relying on, making them practically relevant.\comment{ We exemplify this in the next section.}

\paragraph{Other notions of clustering stability}
 A different notion of clustering stability has been proposed for
 model selection in a line of work including
 \citet{ben-david:06,Rakhlin2006,Shai2007-kmeans,BenDavidBoundary,OhadTishby2009,Shamir2010-MLJournal};
 we will call it {\em output stability} to distinguish it from our
 own stability definition. The aim is to validate a number of
 clusters $K$ by measuring the variability to sampling noise of a
 clustering algorithm with parameter $K$. When the variability of the
 algorithm's output is small then we have output
 stability, and $K$ is ``correct''. \mmp{ framework works directly on
 characterizing the population but not algorithmic result.} An issue with output stability is that, when $\loss(\Pp,\clust)$ has a unique global minimum, output stability is implied for large $n$~\citep{ben-david:06}. Moreover, output stability is implied by $(\Delta, \epsilon)$ stability. Hence, our work provides tractable (but conservative) ways to verify output stability. Furthermore, current output stability results cannot  associate finite sample stability with population ones. One could
 easily construct populations which are asymptotically output stable but with
 arbitrarily large sample size still lack finite sample output stability
 \citep{BenDavidBoundary}. \comment{In our result from theorem
 \ref{thm:instable}, we manage to test population instability from
 finite sample instability with nominal provable probabilistic bounds.}
 \hanyuz{I believe their example shows that asymptotic stability does
   not necessarily leads to finite sample stability in their
   sense. For our result I feel that this is not enough -- to draw the
   conclusion we need to include theorem 7 to better say so...}

\paragraph{Other work in unsupervised learning}
In \cite{hazanMa:unsup16}  a PAC-like
framework for unsupervised learning is proposed. Similar to our paper, the framework of
\cite{hazanMa:unsup16} argues for the need of a {\em hypothesis
  class}, of an assumption that the data {\em fits the model class}
(i.e., the $(k,\epsilon)$ decodability condition), and the use of
problem specific tractable relaxations as vehicles for both tractable
algorithms and error bounds. The difference is that they
concentrate on prediction, not cluster structure. For instance,
under the framework of \cite{hazanMa:unsup16}  one could provide very good guarantees for
 data that is not clusterable, such as the data in Figure 
\ref{fig:example-kmeans}, right.

\section{Experimental evaluations}
\label{sec:experiments}
\subsection{K-means guarantees}
\label{sec:exp-kmeans}
\begin{figure}
\setlength{\picwi}{0.45\textwidth}
\begin{tabular}{cc}
$K=4$, separation $\approx 5.66$ &
$K=6$, separation 1\\
\includegraphics[width=\picwi]{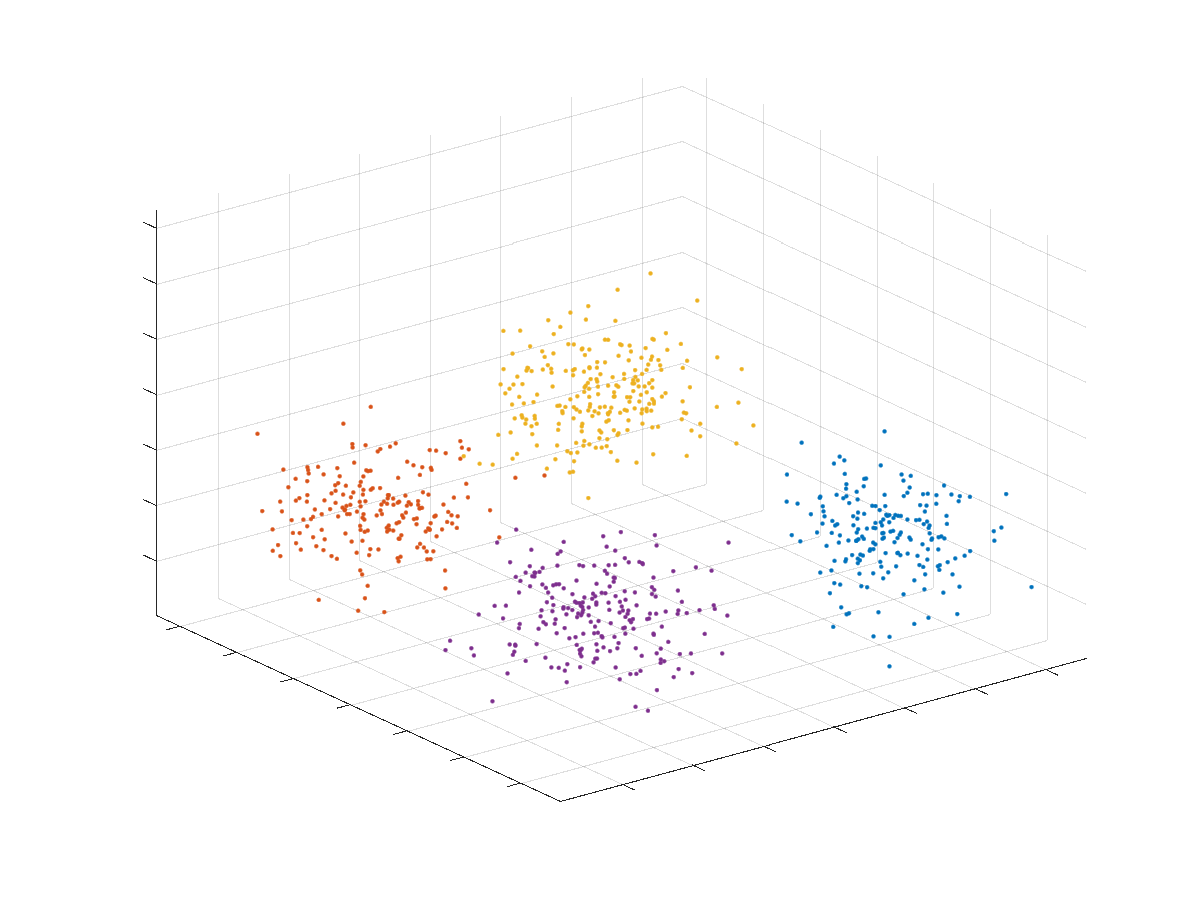}
&
\begin{tabular}[b]{c}
\includegraphics[width=\picwi]{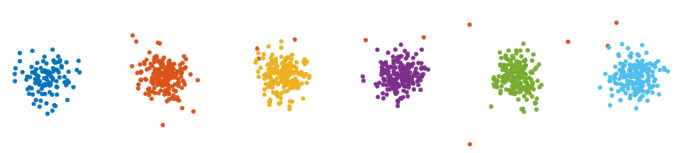}
\\
\includegraphics[width=\picwi]{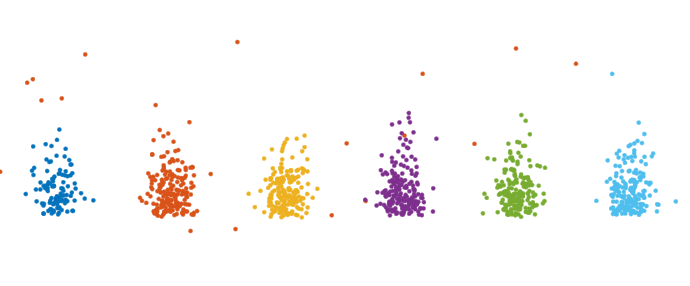}
\\
\end{tabular}
\\
$\sigma=.6$ &
$\sigma=.1$ \\
\includegraphics[width=\picwi]{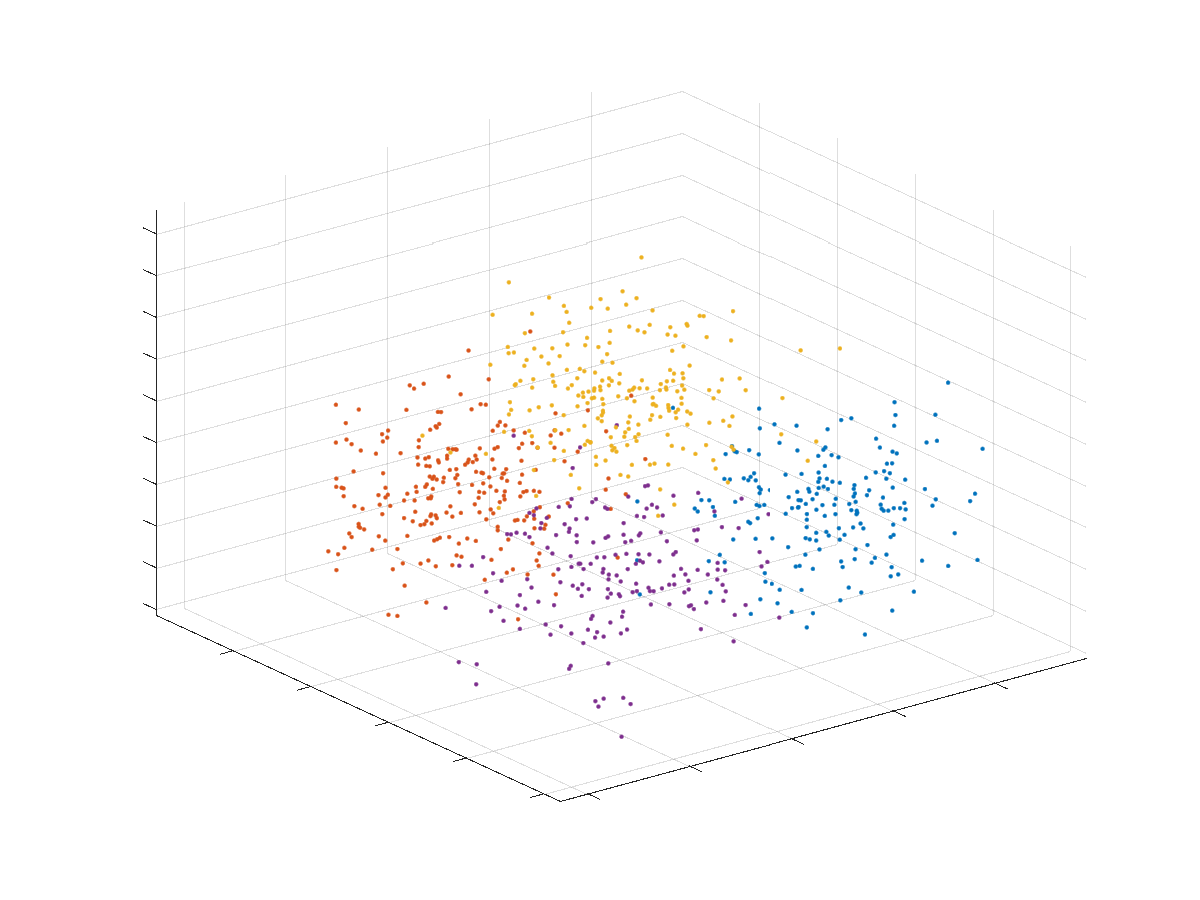}
&
\begin{tabular}[b]{lrrrrrr}
\multicolumn{4}{c}{$K=6$ clusters}\\
\cmidrule{1-4}
&\multicolumn{1}{c}{normal} &&\multicolumn{1}{c}{non-normal}
  \\
 \cmidrule{2-2} \cmidrule{4-4}
  $\sigma$&\multicolumn{1}{c}{$n=525$}&&\multicolumn{1}{c}{$n=525$}
  \\
  \cmidrule{2-2} \cmidrule{4-4}
{0.06}&0.00(0.00)&&0.005(0.001)\\
{0.08}&0.01(0.00)&&0.006(0.001)\\
{ 0.1}&0.01(0.00)&&0.009(0.003)\\
\\
\\
\bottomrule
\end{tabular}
\\
$\sigma=1.1$ 
\\
\end{tabular}
\caption{\label{fig:data}
Some data used in the experiments. In the first three plots, the clusters are sampled from mixtures of spherical Gaussians. In the last, one of the 15 coordinates is from a Gamma$(2,.4)$ distribution and rescaled by $\sigma$. Separation is the distance between the Gaussian means, and $\sigma$ is the standard deviation of the Gaussians. The $K=6$  data sets are designed to be hard for the spectral bounds but not for the SDP bounds. Bottom, left: optimality intervals $\epsi$ for data sampled from normal and non-normal mixtures with $K=6$ (mean and standard deviation over 10 replications). The values of $\epsi_{Sp}$ were much larger than $\pmin$ and are shown.}
\end{figure}

We implemented the $SS\kmeans$ problem using the SDP solver
SDPNAL+\cite{zhaoST:sdpnal10,yangST:sdpnal15}. We also implemented the
spectral bound of \cite{M:kmeans-distortion-icml06}, the only other
method offering optimality intervals for K-means. The main questions
of interest were (1) do our OI exist for realistic situations? (2)
how tight are the bounds obtained? \mmp{and (3) given that SDP solvers are
computationally demanding, can this approach be applied to reasonably
large data sets?}  \mmp{Experiments gave remarkably good answers to the
  first two questions, and showed promise for the third.}
  
  \paragraph{Synthetic Data of Stochastic Ball Model} For this setting we sampled data uniformly from $K$ balls in dimension $d$ with unit radius. Let $\Delta$ be the minimal distance between the centers of the $K$ balls, then the SDP relaxation of K-Means we adopted in this paper can guarantee the exact recovery of $K$ clusters with high probability when $\Delta > 2+\epsilon(d)$, where $\epsilon(d)\rightarrow 0$ as $d\rightarrow \infty$. This means that under this specific stochastic ball model, the k-means guarantee can only be obtained when the seperation between balls are large enough so that they don't touch each other. 
 
\begin{figure}
 \centering
 \includegraphics[width = 0.5\textwidth]{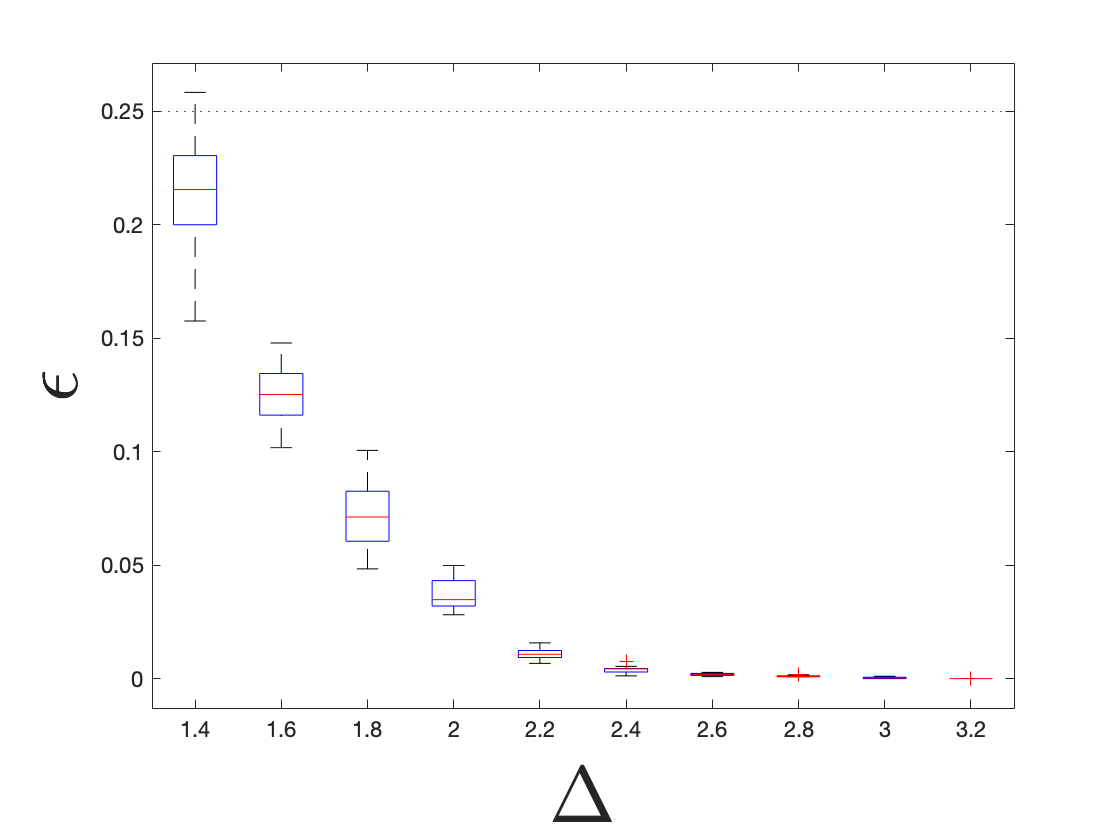}
 \caption{\small \label{fig:sbm}
 $\epsilon$ bound obtained for data sampled from stochastic ball model. The procedure is repeated for $10$ times under each setting. The $\epsilon$ bound are meaningful when they are smaller than $p_{\min}\approx 0.25$. }
  \end{figure}

 We sampled $n=500$ data from the stochastic ball model with $K=4,d=2$ and $\Delta$ ranging from 1.4 - 3.2. The centers of each ball are aligned on one line segment with equal space between. Then we perform K-means clustering with the initialization of correct labels, since we are only interested in understanding the behaviour of our method's ability to obtain a guarantee for clustering result. Under this setting, the theoretical bound is trivial (larger than 4). Theoretically we can say nothing about how good the SDP relaxation is for the clustering. However as shown in figure \ref{fig:sbm} our method can provide some guarantees on a particular clustering result. In all settings, initializing with the true label, the distance between true label and the K-means solutions are close, with earth mover distance approximately $\sim 10^{-6}$. We see that in the model touching cases, our SS method provides insights on the distance between the K-means global optimal solution to the true label. On the other hand, all $\Delta \geq 2.2$ are within the regime where previous theoretical guarantee does not work. The results shows that  K-means global optimizers approximately achieve the exact recovery of  stochastic ball models.
 
\paragraph{Synthetic Data of Gaussian Mixture} We sampled data from a mixture of $K=4$ normal distributions with equal spherical covariances $\sigma^2 I_d$, in $d=15$ dimensions. The cluster sizes $n_k$ were approximately equal to $\lfloor n/K\rfloor$. The cluster means were at the corners of a regular tetrahedron with center separation $||\mu_k-\mu_{k'}||=4\sqrt{2}\approx 5.67$.  The data was clustered by K-means with
random initialization, then the bounds $\epsi,\epsi_{Sp}$ corresponding respectively to the SS method and to the spectral method of \cite{M:kmeans-distortion-icml06} were computed. 
\begin{figure}
\setlength{\picwi}{0.5\textwidth}

\begin{tabular}{ccc}
$n=256$ & $n=1024$\\ 
\includegraphics[width=\picwi]{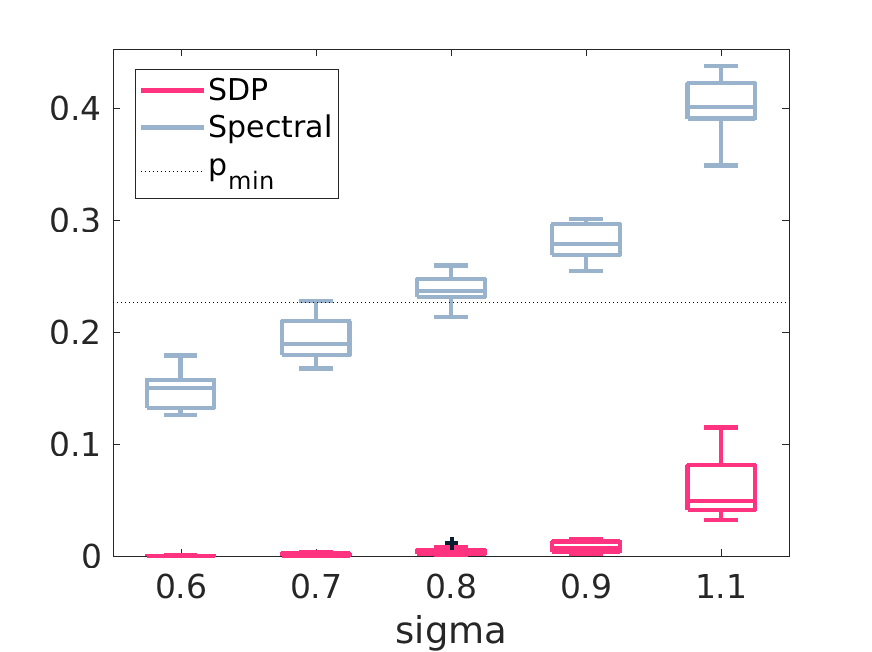}
&
\includegraphics[width=\picwi]{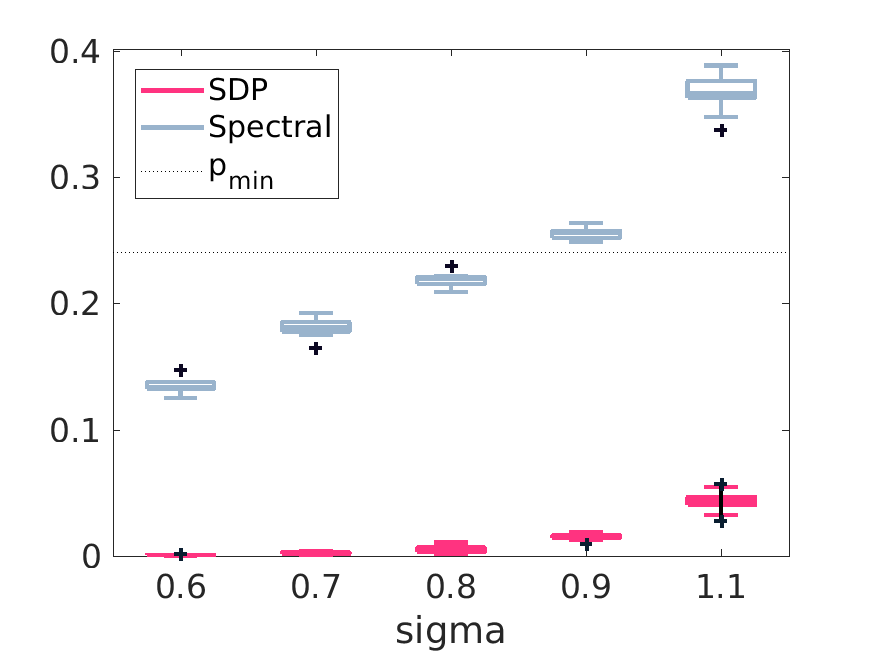}\\
\end{tabular}
\caption{\small \label{fig:k4}
The optimality intervals $\epsi$ and $\epsi_{Sp}$ for data sampled from mixtures of normal distributions with $K=4$, $n=256$ and $1024$ and various $\sigma$ values (over 10 replications). The values of $\epsi_{Sp}$ exceeding $\pmin$ are not valid. Note that the $\epsi$ bounds are near 0 even though the clusters are not separated for $\sigma>0.8$.
}
\end{figure}
In the experiments we also performed {\em outlier removal}, as follows. For each $x_i$, we computed the sum of the distances to its $\pmin/2$ nearest neigbors. We then removed the $n_0$ data points with the largest values for this sum. For good measure, we first added 20 outliers, then removed $n_0=4\%n$ respectively $n_0=2\%n$  points  (so that $n_0$ is slightly larger than 20), before computing the bounds $\epsi,\epsi_{Sp}$. Consequently, these bounds do not refer to clusterings of the original $\dataset$, but to the ``cleaned'' dataset. Note that the outlier removal does not depend on the cluster labels; it is performed before clustering the data.
\begin{figure}
\setlength{\picwi}{0.5\textwidth}
\begin{tabular}{cc}

\includegraphics[width=\picwi]{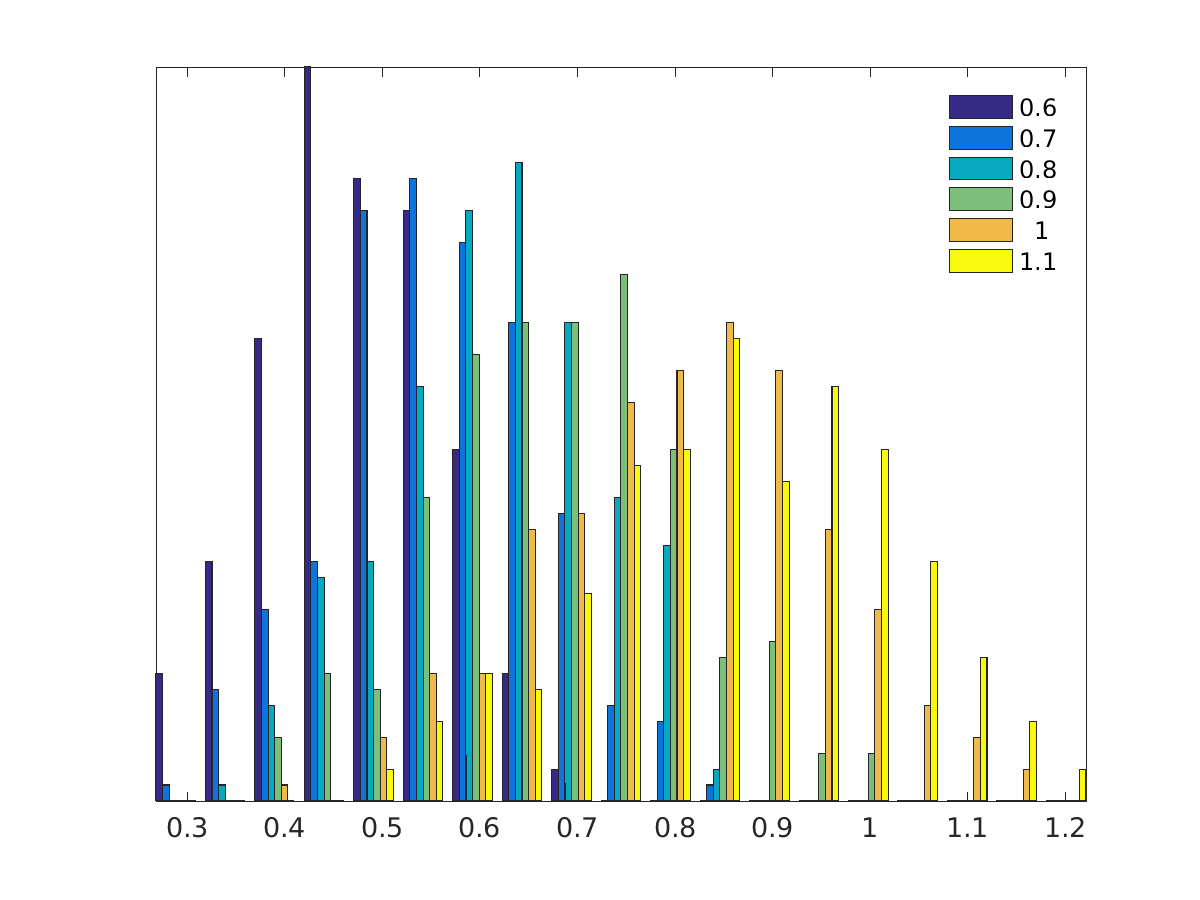}
&
\includegraphics[width=\picwi]{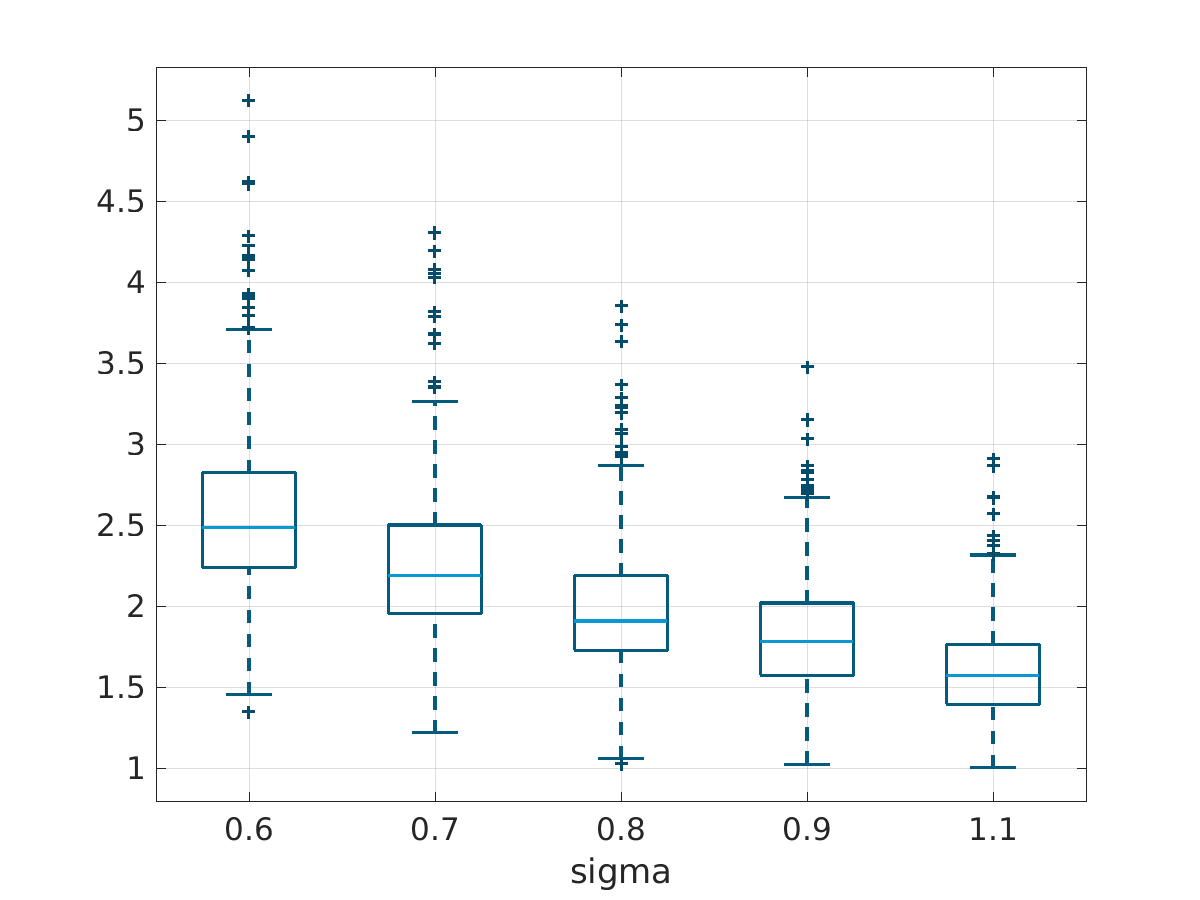}\\
\end{tabular}
\caption{\label{fig:sepk4}
  Separation statistics for the $K=4$ data, $n=1024$, all $\sigma$ values. Left: histogram of $\min_k||x_i-\mu_k||/\min_{k,k'}||\mu_k-\mu_{k'}||$ (i.e. distance of point to its center over minimum center separation) colored by $\sigma$. Note that when the clusters are contained in equal non-intersecting balls this ratio is strictly smaller than $0.5$. Right: boxplot of distance to second closest center over distance to own center, versus $\sigma$.  }
\end{figure}
Figure \ref{fig:k4} displays the bounds $\epsi,\epsi_{Sp}$ for these data, while Figure \ref{fig:data}, top and bottom, left, displays some representative samples. The $\epsi$ optimality interval is much tighter than the spectral one
$\epsi_{Sp}$, and, surprisingly enough, holds even when the clusters
``touch'', i.e when there is no region of low density between the
clusters.  Figure
\ref{fig:sepk4} (left) shows that, when $\sigma > 0.8$, the minimal
spheres containing the clusters intersect; on the right we see that
there are points which are almost equidistant from two cluster
centers.\comment{, and there are point $i,j$ which belong to different
  clusters but are close to each other.}\comment{It is remarkable too,
  and it surprised us, that we did not need to carefully seed K-means
  to obtain obviously good clusterings in these cases. This suggests
  that the current algorithmic results such as \cite{awasthiBV:icml14}
  are too conservative.}
Otherwise put, the {\em distribution free} bounds hold
even when the data are not contained in non-intersecting balls, which
is the best known condition for clusterability {\em under model
  assumptions} \cite{Aswathi+Ward:14,awasthiBV:icml14}.
\comment{Table \ref{tab:bound-ksimplex-unequal} shows the results for $K=6$, $n=XX$; the spectral bound $\epsi_{Sp}$ was never valid for $K=6$ and is therefore not shown. }

Next, we performed experiments with unequal cluster sizes
$p_{1:4}=0.1,0.2,0.3,0.4$, and we also
generated non-gaussian clusters (details in the Appendix). We also performed experiments with $K=6$ clusters, with $p_{1:6}=0.1,0.18,0.18,0.18,0.18,0.18$. For $K=6$ we placed cluster centers along a line, as shown in Figure \ref{fig:data}, top and bottom, right. This hurts the spectral bound which depends on a stable $K-1$-subspace, but does not hurt, and may even help the SDP bound $\epsi$. 

The results are shown in Table \ref{tab:bound-ksimplex-unequal}, with run times for all experiments in the Appendix. The spectral bound $\epsi_{Sp}$ was much larger than $\epsi$ for all $K14$ experiments, and was never valid for $K=6$; therefore, it is omitted. The table also shows that $\epsi$ takes similar values in the case of equal and unequal
clusters. However, in the latter case, the condition $\epsi\leq\pmin/\pmax$, is more stringent, hence some of the bounds obtained are
not valid.

Over all experiments, we have found that the values of $\epsi$ are
virtually insensitive to the sample size $n$, and degrade slowly when
$\pmin$ decreases. The main limitation in these experiments is the
requirement that $\epsi\leq \pmin$. This requirement can be traced
back to Theorem \ref{thm:equivalence-ML10}.  The bound in this
Theorem, even though state of the art, is not tight, suggesting that some the $\epsi$ values marked in gray are valid OI even though we cannot prove it at this time.
\begin{table}
\caption{\label{tab:bound-ksimplex-unequal}
The OI $\epsi$ for $K=4$ clusters of unequal sizes (mean and standard deviation over 10 replications). The values in \textcolor{mygray}{gray} are not valid, owing to the fact that $\epsi\pmax>\pmin$ in these cases. Bounds for smaller $\sigma$ values were essentially zero and are ommitted.}
%\centering%\centerline{$K=4$}
\begin{tabular}{lrrrrrr}
  \toprule
&\multicolumn{3}{c}{Unequal normal clusters}
&\multicolumn{3}{c}{Unequal non-normal clusters}\\
  \cmidrule{2-4}  \cmidrule{5-7}
  $\sigma$&\multicolumn{1}{c}{$n=200$}&\multicolumn{1}{c}{$n=400$}&\multicolumn{1}{c}{$n=800$}&
  \multicolumn{1}{c}{$n=200$}&\multicolumn{1}{c}{$n=400$}&\multicolumn{1}{c}{$n=800$}\\
 \cmidrule{2-4}  \cmidrule{5-7}
{0.6}&0.00(0.00)& 0.00(0.00)& 0.00(0.00)  &0.001(0.001)&0.001(0.000)&0.002(0.007)\\ 
{0.8}&0.01(0.01)& 0.01(0.01)& 0.01(0.01)  &0.006(0.004)&0.004(0.002)&0.007(0.003)\\ 
{1.0}&\textcolor{mygray}{0.09 }\textcolor{mygray}{(0.05) }& \textcolor{mygray}{0.06 }\textcolor{mygray}{(0.01)}& \textcolor{mygray}{0.07 }\textcolor{mygray}{(0.02) }
&{0.04 (0.02) }&{0.03 (0.01) }&{0.03 (0.01) }\\       
{1.2}&\textcolor{mygray}{0.28 }\textcolor{mygray}{(0.08) }& \textcolor{mygray}{0.21 }\textcolor{mygray}{(0.05) }& \textcolor{mygray}{0.21 }\textcolor{mygray}{(0.03) }
&\textcolor{mygray}{0.16 }\textcolor{mygray}{(0.06) }&\textcolor{mygray}{0.14 }\textcolor{mygray}{(0.03) }&\textcolor{mygray}{0.13 }\textcolor{mygray}{(0.03) }\\       
\bottomrule
\end{tabular}
\comment{ksimplex, unequal, from Figures-july17 (table with plot\_sdp\_results.m, separately for each n, merged, edited by hand afterwards)
(To redo equal clusters?)}
\end{table}

\comment{ % RUN TIMES to still put in

{\bf sixpack experiments march 2018 $n=525$ small pmin}
Run times
\begin{tabular}{|l|r|r|}
\hline
\textbf{0.06}&560.49(395.78)\\\hline
\textbf{0.08}&777.11(264.16)\\\hline
\textbf{ 0.1}&741.62(465.90)\\\hline
\textbf{0.12}&1294.76(466.11)\\\hline
\end{tabular}

{\bf sixpack experiments march 2018 $n=525$ pmin $= 0.1$}
\\
\begin{tabular}{|l|r|r|}
\hline
\textbf{0.06}&0.00(0.00)\\\hline
\textbf{0.08}&0.01(0.00)\\\hline
\textbf{ 0.1}&0.01(0.00)\\\hline
\end{tabular}
Run times
\begin{tabular}{|l|r|r|}
\hline
\textbf{0.06}&238.17(122.32)\\\hline
\textbf{0.08}&374.98(181.25)\\\hline
\textbf{ 0.1}&720.63(376.13)\\\hline
\end{tabular}
\\
(all valid!) to also run 0.12
}
\comment{
\begin{tabular}{lllllll}
\toprule
&\multicolumn{3}{c}{$K=6,\pmin\approx0.16$, center sep$\approx 1$} && 
% empty column to separat sixpack from octoclus
\multicolumn{2}{c}{$K=8,\pmin\approx0.12$, center sep$\approx 1$} \\
\cmidrule{2-4}  \cmidrule{6-7}
&$n=192$ & $n=384$ &$n=768$ && $n=256$ & $n=1024$ \\
\midrule
$\sigma=.1$ &.01(.005)& .01(.006) &  .01(.004) &&
.07($\approx 0$)&   .003(.006)\\
$\sigma=.13$ &.02(.011)& .018(.007)& .02(.011) &&
-- & .012(.01)\\ %in reality these are for sigm=0.14 or .15
%$\sigma=.15$ &.03(.01)& & & &
\midrule
time[s] & 200 & 550 & 2,130 && 9 &6,000\\
\bottomrule
\end{tabular}
}% end comment
\comment{
  pmin, runtime mean and median for sixpack (see junk1.m)
  .1667    .1635    .1635    .1667    .1646    .1667    .1655
 1.0e+03 *
  .0876    .1291    .1571    .4016    .5656    1.7246    2.3973
 1.0e+03 *
  .0795    .1362    .2018    .3402    .5589    1.5974    2.1292
}

\begin{figure}
\setlength{\picwi}{0.45\textwidth}
\begin{tabular}{lll}
\hspace{-1em}\includegraphics[height=0.3\picwi]{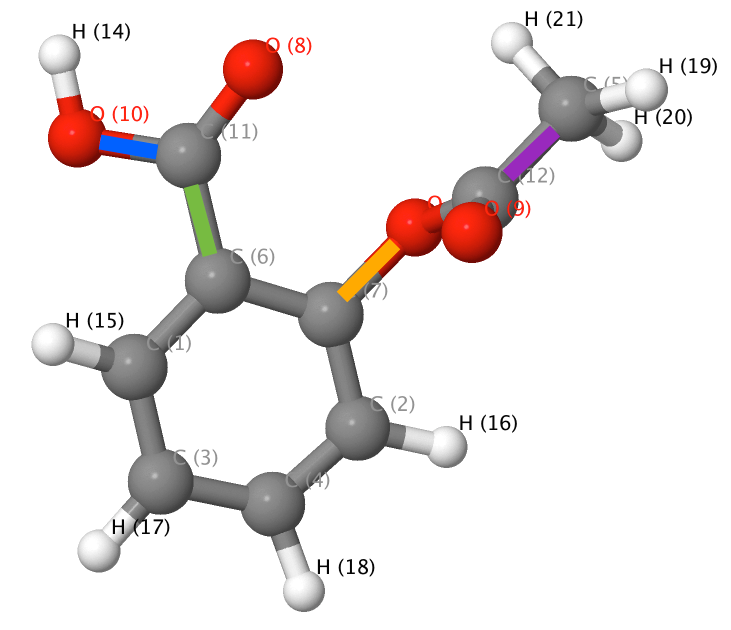}
&
\includegraphics[width=0.3\picwi,angle=90]{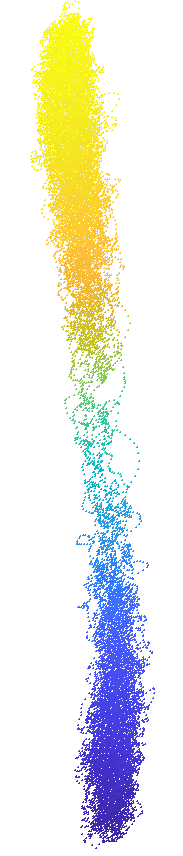}
&
\begin{minipage}[b]{0.35\picwi}
\small
$K=2$\\$\pmin=.26$\\$\pmax=.74$\\
\end{minipage}
\end{tabular}
\caption{\label{fig:aspirin}
Left: the aspirin molecule. Middle: The first two principal components of the 57-dimensional aspirin data. The data is a Molecular Dynamics sequence of 211,762 configurations. We sample every 100-th point of the data for clustering. The axes are represented {\em at scale}.
}
\end{figure}

\comment{
mean distance to centers for K=4
  2.2665
    2.6420
    3.0176
    3.3752
    4.1419
mean center separation 5.65, min\approx5.3}
\comment{ octoclus (see junk1)

%   1.0e-03 *
%    0.6541    0.0000         0    0.0000         0
%    0.1250    0.1240    0.1250    0.1250    0.1250
%   1.0e+03 *
%    1.8328    6.0553    0.0195    0.0675    0.0089
%   1.0e+03 *
%    1.3033    6.0553    0.0195    0.0675    0.0089
}
\comment{ % Sixpack equal clusters -- SHALL I PUT SOME OF THIS IN??
\begin{tabular}{lllllll}
\toprule
&\multicolumn{3}{c}{$K=6,\pmin\approx0.16$, center sep$\approx 1$} && 
% empty column to separat sixpack from octoclus
\multicolumn{2}{c}{$K=8,\pmin\approx0.12$, center sep$\approx 1$} \\
\cmidrule{2-4}  \cmidrule{6-7}
&$n=192$ & $n=384$ &$n=768$ && $n=256$ & $n=1024$ \\
\midrule
$\sigma=.08$ &.01(.005)& .01(.006) &  .01(.004) &&
.07($\approx 0$)&   .003(.006)\\
$\sigma=.10$ &.02(.011)& .018(.007)& .02(.011) &&\\
$\sigma=.12$ &.03(.01)& & & &\\
$\sigma=.14$ &.03(.01)& & & &\\
\midrule
time[s] & 200 & 550 & 2,130 && 9 &6,000\\
\bottomrule
\end{tabular}
}
\comment{% Now most of this is in Table 3
\begin{tabular}{lllllll}
\toprule
&\multicolumn{3}{c}{$K=6,\pmin\approx0.044$, center sep$\approx 1$} && 
% empty column to separat sixpack from octoclus
\multicolumn{2}{c}{$K=8,\pmin\approx0.12$, center sep$\approx 1$} \\
\cmidrule{2-4}  \cmidrule{6-7}
& $n=525$ &$n=1050$ && $n=540$ & $n=1080$ \\
\midrule
$\sigma=.08$ && \textcolor{mygray}{0.02(0.00)}&  && & \\
$\sigma=.10$ && \textcolor{mygray}{0.02(0.01)}&  &&\\
$\sigma=.12$ && \textcolor{mygray}{0.04(0.01)}&  &&\\
\midrule
time[s] &   & 6,563(1,490) &&  &\\
\bottomrule
\end{tabular}
}
%
%Above, as well as in the other experiments, we observed that $\epsi,\epsi_{Sp}$ were almost independent of the sample size $n$. 
%
\paragraph{Real data: configurations of the aspirin ($C_9H_8O_4$) molecule}
\label{sec:exp-real}
These $n=2118$ samples (see Figure \ref{fig:aspirin}) were obtained
via Molecular Dynamics (MD) simulation at $T=500$ degrees Kelvin by
\cite{chmielaTkSauPSchM:energy17} and represent 3D positions of the 21
atoms of aspirin. It was discovered recently that aspirin's potential
energy surface has two energy wells. The purpose of clustering of MD
simulations is to label the data by energy well; this allows chemists
to identify energy wells, on one hand, and on the other to identify
states around on the transition path between the energy wells. These
states describe the mechanism of a transition, and, being rare events
in the contex of many simulations, are of great interest.  Currently,
the labeling is done by ad~hoc algorithms. Having guarantees of
(almost) correctness for the grouping, such as those in Figure
\ref{fig:chrisfu-bounds} saves the time needed to validate the
clustering by human inspection. Hence, we cluster these data into $K=2$ clusters, after having removed $n_0=0.5\%n=106$ outliers. The clusters found have relative sizes $\pmin=.26,\pmax=.74$, and the OI is $\epsi=.065$, an informative bound. \comment{However, this took over 10h\comment{17h}; that encouraged us to try the following heuristic: instead of removing outliers, we removed 60\% of the data points, which were {\em closest to their centers}. The motivation was that the difficulty of the SDP depends on the cluster boundaries and not on the easy points. The run time reduced to 42 minutes and the bound obtained $\epsi=0.047$ was comparable with the original one. \comment{847 points +1271 central points removed = 2118=41 min; bound = 0.11 * (847/2118)=0.047;; 106 outliers + 2012 points = 2118 = 17.5h} While this speed-up method is ad-hoc, we are confident that it can be made rigurous in the future, opening up the SS method to larger data sets.}\comment{
shows that using SDP solvers for SS problems can become practical,\comment{and further on we enumerate a few possible approaches.}}
\comment{TABLE WITH SIXPACK
\begin{center}
  $K=6$ clusters\\
\begin{tabular}{lrrrrrr}
  \toprule
&\multicolumn{2}{c}{$p_k\propto k$, $\pmin\approx0.045$}
  &&\multicolumn{1}{c}{$\pmin\approx0.1$}
  &&\multicolumn{1}{c}{$\pmin\approx0.1$}
  \\
  \cmidrule{2-3}  \cmidrule{5-5} \cmidrule{7-7}
  $\sigma$&\multicolumn{1}{c}{$n=525$}&\multicolumn{1}{c}{$n=1050$}&
   &\multicolumn{1}{c}{$n=525$}&&\multicolumn{1}{c}{$n=525$}%&\multicolumn{1}{c}{$n=800$}
  \\
 \cmidrule{2-3}  \cmidrule{5-5} \cmidrule{7-7}
{0.06}&\textcolor{mygray}{0.02(0.00)}&0.02(0.00) &&0.00(0.00)&&0.005(0.001)\\
{0.08}&\textcolor{mygray}{0.02(0.01)}&0.02(0.00) &&0.01(0.00)&&0.006(0.001)\\
{ 0.1}&\textcolor{mygray}{0.03(0.01)}&0.02(0.01) &&0.01(0.00)&&0.009(0.003)\\
{0.12}&\textcolor{mygray}{0.04(0.01)}&0.04(0.01) && && non-normal\\
\bottomrule
\end{tabular}
\end{center}
\comment{\label{tab:bound-sixpack-unequal}BTR bound $\epsi$ for $K=6$ clusters of unequal sizes (mean and standard deviation over 10 replications). The values in \textcolor{mygray}{gray} are not valid, owing to the fact that $\epsi\pmax>\pmin$ in these cases. Bounds for smaller $\sigma$ values were almost all 0. Center separation $\min_{k,k'}||\mu_k-\mu_{k'}||$ is $5.67$ for $K=4$ and 1 for $K=6$.
}%end comment/end caption
\comment{sixpack-out-n1045-k6, unequal, from Figures-july17. 1050,0.06 not found; to redo! }
}%end comment
\subsection{Normalized Cut guarantees}
\label{sec:experiments-ncut}
\begin{figure}[t]
\setlength{\picwi}{0.3\textwidth}
\begin{tabular}{ccc}
$S$ for $\sigma=32$ & degrees $D_{ii}$,$\sigma=2$ & (log) degrees $\sigma=2$ \\ 
\includegraphics[width=\picwi]{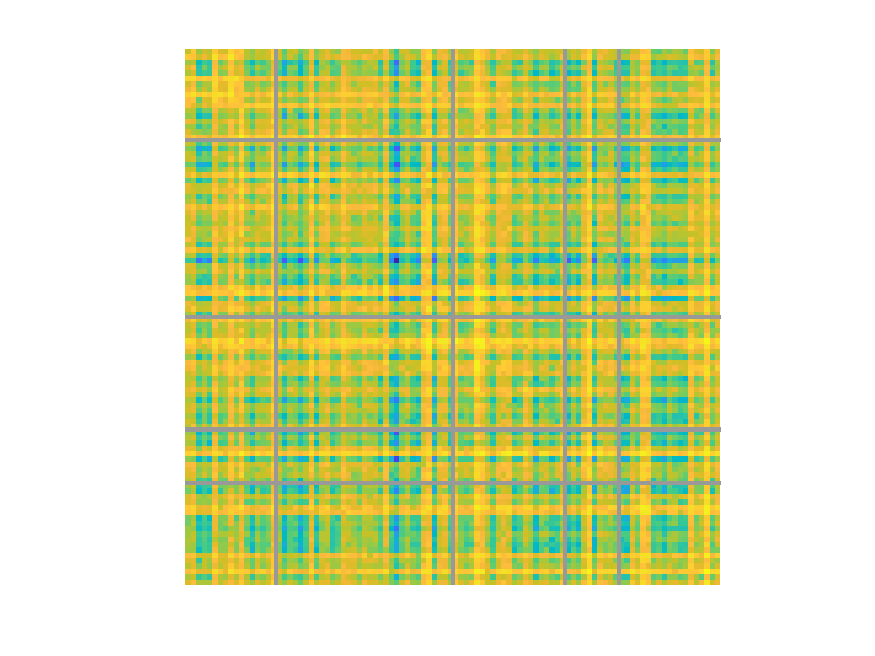}
&
\includegraphics[width=\picwi]{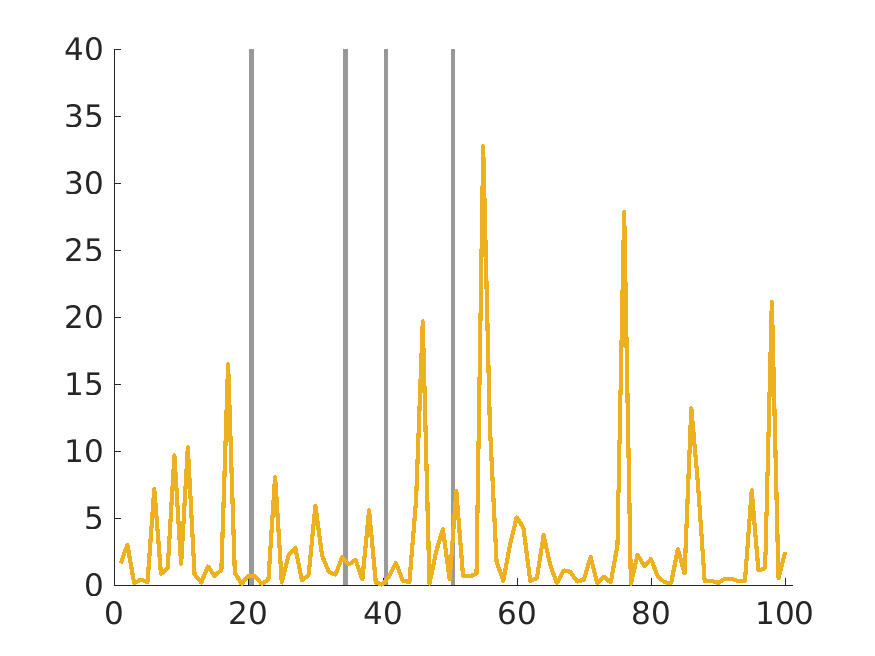}
&
\includegraphics[width=\picwi]{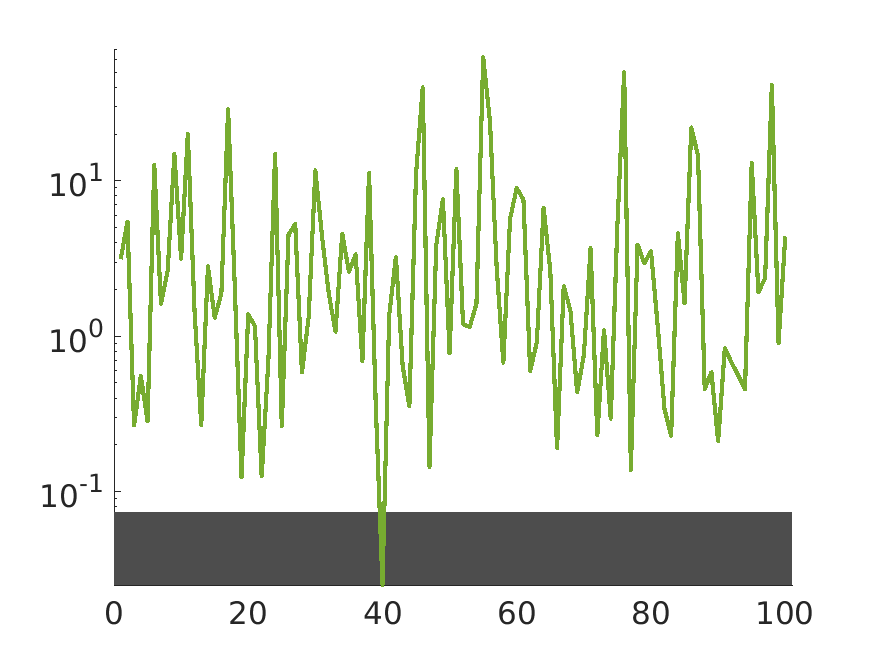}
\\
eigenvectors of $L$, $\sigma=32$ & degrees $D_{ii}$, $\sigma=32$ & (log) degrees $\sigma=32$ \\ 
\includegraphics[width=\picwi]{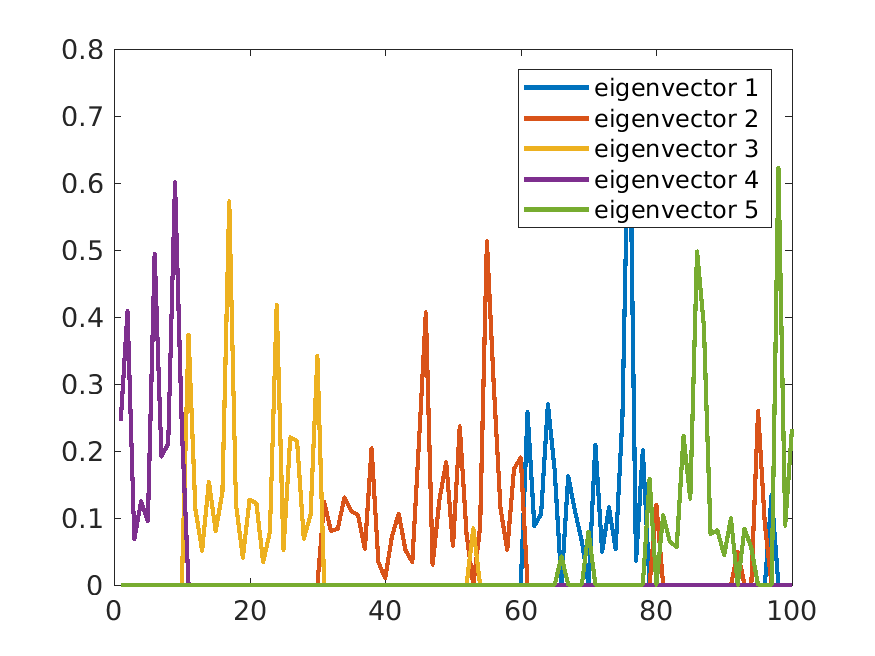}
&
\includegraphics[width=\picwi]{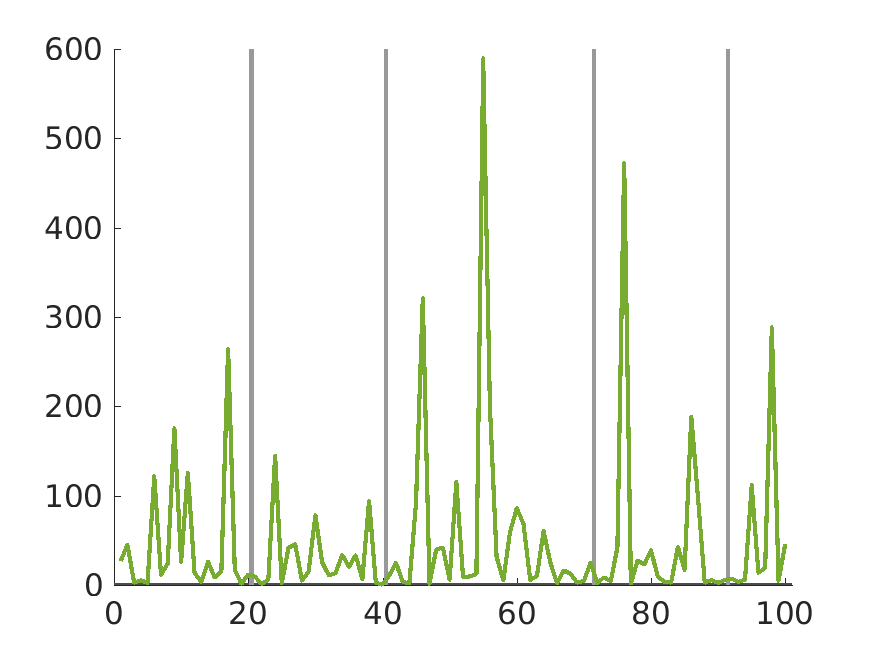}
&
\includegraphics[width=\picwi]{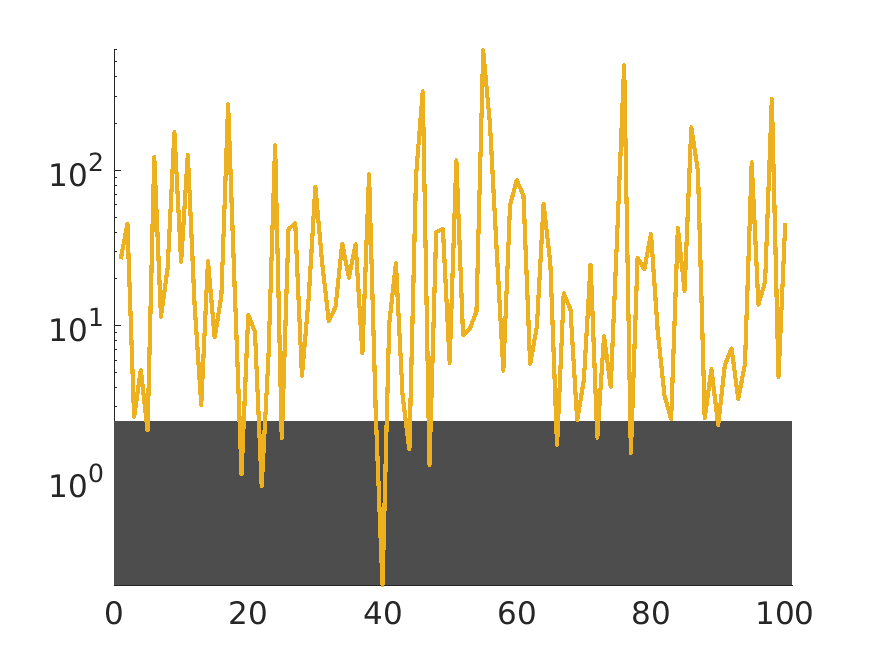}
\\
\end{tabular}
\caption{\label{fig:ncut-synth-data} Illustrating the properties of the synthetic similarity matrices used in experiments; $K=5$, $n=100$, noise of amplitude $\sigma$.  Example of similarity matrix $S$ {\bf (top, left)} (values are on a logarithmic scale for better visualization); principal $K$ eigenvectors of the corresponding $L$ {\bf (bottom, left)}; values of node degrees $D_{ii}$ for $\sigma=2$ {\bf (top, middle)} and $\sigma=32$ {\bf (bottom, middle)} (note that the degrees are approximately proportional to $1+\sigma$); same node degrees on a logarithmic scale, with dark gray area representing the level of the SS bound $\epsi$ obtained for these data {\bf (top and bottom, right)}. Nodes with degrees above $\epsi$ are guaranteed not to change cluster membership in any clustering as good as the one found.
  }
\end{figure}

\begin{figure}[t]
\setlength{\picwi}{0.5\textwidth}
\begin{center}
OI $\epsi$ and $\epsi_{Sp}$\\% & Run time [s]\\
\includegraphics[width=\picwi]{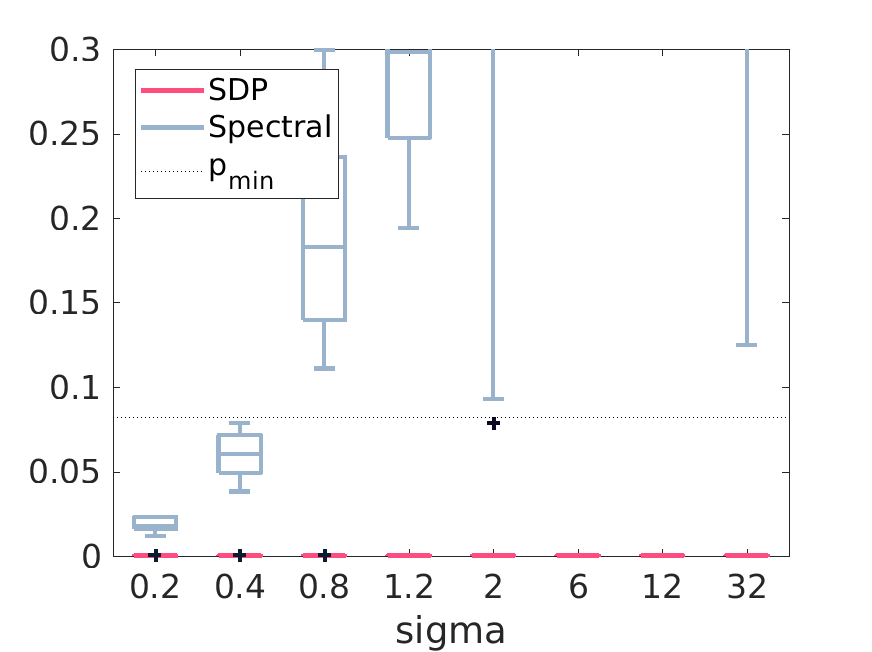}
\end{center}
%TODO: what to do if pmin = 0? better declare no bound. to redo experiments
\caption{\label{fig:ncut-synth} Experimental results for synthetic similarity matrix, clustered by Spectral Clustering; $K=5$, $n=100$, noise of amplitude $\sigma$; boxplots of OI $\epsi$ and spectral OI $\epsi_{Sp}$ versus noise amplitude $\sigma$ , over 10 replications for each noise level. Note that the OI from the SS method are practically 0.} 
\end{figure}

\paragraph{Synthetic data}
The matrix $S$ shown in Figure \ref{fig:ncut-synth} was generated according to \cite{MShi:nips00}. Even though $S$ is not block diagonal, it admits a perfect clustering, in the sense that the $K$ principal eigenvectors of $D^{-1}S$ are constant over each cluster. We perturbed $S$ with non-negative noise by $S_{ij}\leftarrow S_{ij}(1+\sigma u_{ij})$, $S_{ji}\leftarrow S_{ij}$, where $u_{ij}\sim uniform[0,1]$, i.i.d. for $i\leq j$ and $\sigma >0$. This perturbation keeps $S$ symmetric and with non-negative elements, but affects the eigenvectors and consequently the clustering, as shown in Figure \ref{fig:ncut-synth-data}.

We obtained a clustering of $S$ by spectral clustering \cite{} with $K=5$, and repeated the process for different noise amplitudes $\sigma$ and random noise realizations. The results are displayed in Figure \ref{fig:ncut-synth}.

\hanyuz{Is this part still in the paper? I did not find the references and the figures.}
\paragraph{Real data: MD simulation of a reversible reaction}
In this section we consider molecular dynamics simulations \citep{flemingTPfae:16} of the reversible reaction 
%\beq \label{eq:reaction}
$CH_3C\ell + C\ell^-\;\leftrightarrow\;CH_3C\ell + C\ell^-,$
%\nonumber
%\eeq
in which one of the chlorine ($C\ell$) atoms replaces the other in the methilchloride molecule. Molecular simulations of a reversible chemical reaction 
typically exhibit two clusters, one for each state of the system. Because of the symmetry of the reaction, the cluster sizes are approximately equal. The
density between the two clusters, where the states {\em on the reaction path} lie, depends on the absolute temperature $T$ of the system. The intercluster density will be lower an the data more clusterable at lower temperatures. The data used in our experiments, available at \url{https://www.stat.washington.edu/spectral/data/MDsimulations2017/}, consist of 10 independently simulated trajectories at each of the four temperatures $T \in \{ 600, 900, 1050, 1200\}$ degrees Kelvin. Trajectories were decimated to create data sets of $n\approx 1000$ points.

\begin{figure}
\setlength{\picwi}{0.3\textwidth}
\begin{center}
  \includegraphics[width=\picwi]{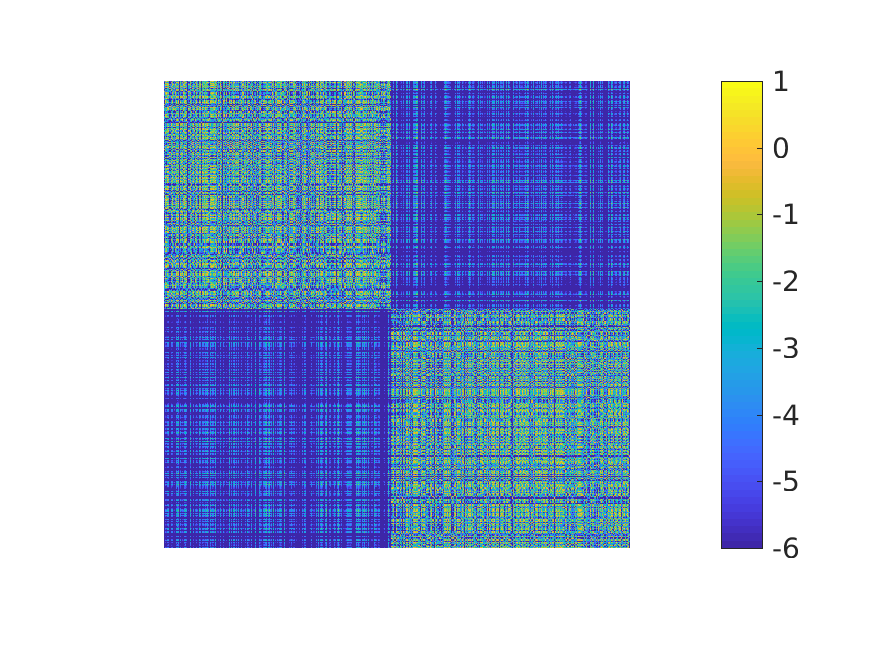}
\includegraphics[width=\picwi]{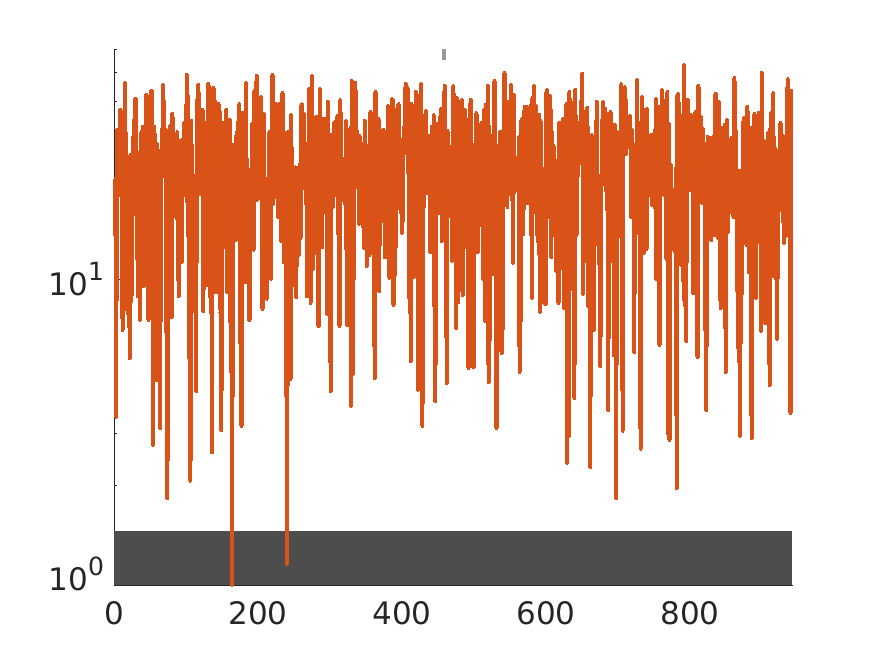}
\includegraphics[width=\picwi]{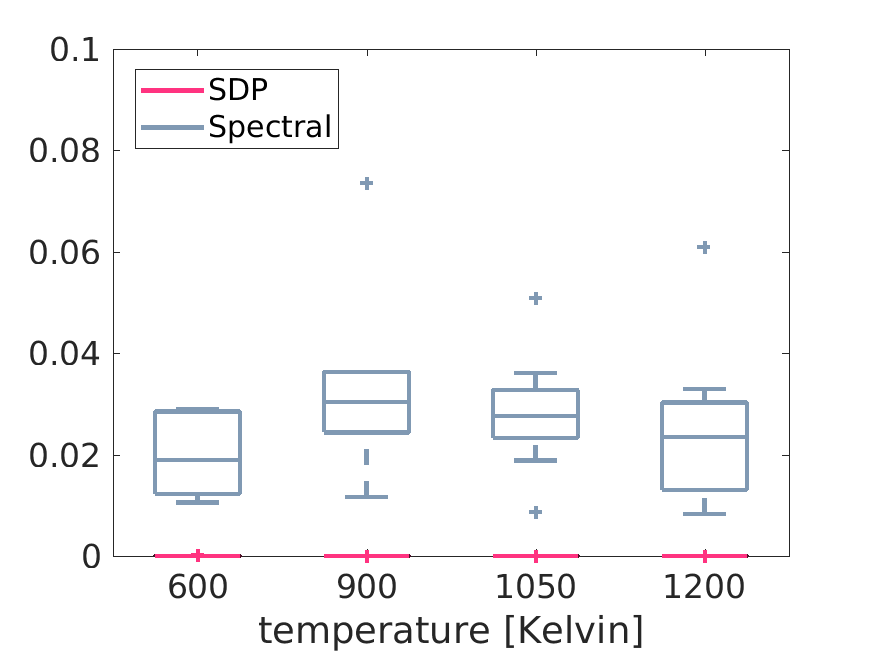}
\end{center}
\caption{\label{fig:chloromet-S}\label{fig:chrisfu-bounds}
Results from the cholorometane experiments; $K=2$, $n\approx 950$, $\pmin\approx 0.47$. Example of similarity matrix $S$ from one simulation at 1200 Kelvin (on a logarithmic scale for better visualization) {\bf (left)}. Values of node degrees $D_{ii}$ for this $S$ matrix, with dark gray area representing the level of the OI $\epsi$ {\bf (middle)}. Nodes with degrees above $\epsi$ are guaranteed not to change cluster membership in any clustering as good as the one found.  The bounds $\epsi$ and $\epsi_{Sp}$ versus temperature {\bf (right)}; there are 10 independent simulations for each temperature. Note that the optimality intervals from the SS method, $\epsi$, are indistinguishable from 0 in this plot. The values of $\pmin$ were all near $0.47$ and are not shown.}
\end{figure}

Figure~\ref{fig:chloromet-S}, displays one of the similarity matrices at the highest temperature (with the data sorted by cluster label) and shows that the node degrees vary by about 2 orders of magnitude.  The OI from Theorem~\ref{prop:ncut} are represented in Figure~\ref{fig:chloromet-S} and summarized below.
\comment{
obtained from The simulation output represents a sequence of $x,y,z$
spatial locations of the six atoms involved in the reaction at
consecutive time steps during the simulation. When a chlorine atom
binds to the methil ($CH_3$) group, the energy of the system is lower;
in the sequence of configurations in which the reaction takes place,
and both chlorine atoms are dissociated, the energy of the system is
higher, hence fewer configurations will be present in these

Because the energy of the molecule depends only on the relative
positions of the atoms, the original $6\times 3=18$ dimensions are
reduced to $12$ degrees of freedom in the following way. First, all
the plane angles within the molecule are computed, for a total of
$\binom{6}{3}=20$ angles. Then, the linear relationships between these are
eliminated by Principal Component Analysis, keeping up to 12
components, or enough to reduce the residual variance to
$10^{-4}$. With the present data, the resulting dimension was always
12. These data {are} now clustered by K-means with $K=2$. Figure
\ref{fig:chrisfu-bounds} shows that the data distribution in each
cluster is non-Gaussian, non-symmetric around the centroids, and
heavy-tailed.  }

\comment{
\begin{figure}
  \setlength{\picwi}{0.5\textwidth}
  \begin{center}
\begin{tabular}{cc}
\includegraphics[width=\picwi]{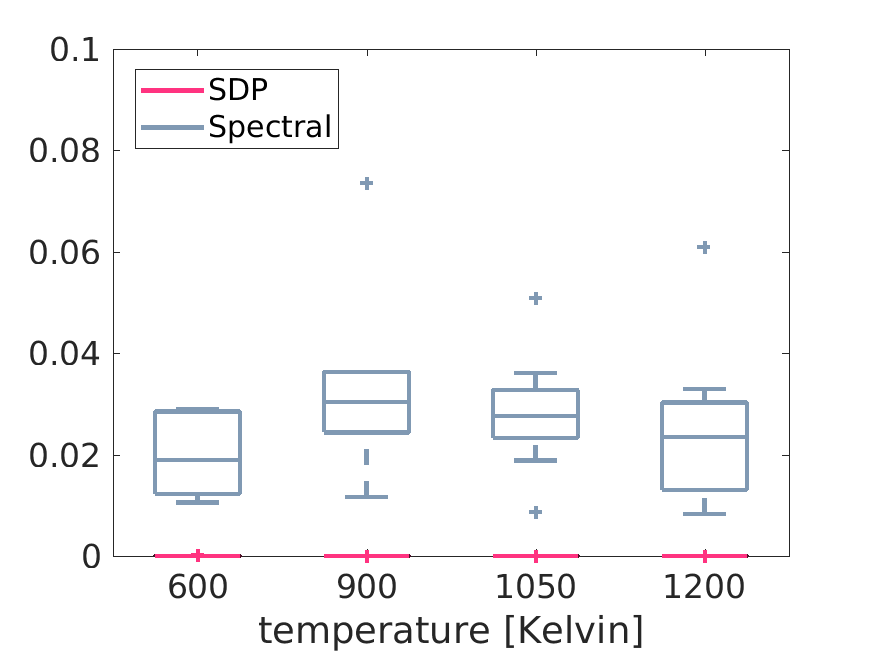}
%\begin{minipage}{0.45\textwidth}
%\end{minipage}
\end{tabular}
\end{center}
\caption{\label{fig:chloromet-eps}
 The bounds $\epsi$ and $\epsi_{Sp}$ versus temperature; there are 10 independent simulations for each temperature. Note that the optimality intervals from the SS method, $\epsi$, are indistinguishable from 0 in this plot. The values of $\pmin$ were all near $0.47$ and are not shown.
}
\end{figure}
}% end comment
%For completeness, the values of $\epsi$ are summarized below.
\begin{center}
\begin{tabular}{rrrrr}
{\bf temperature [Kelvin]} & 600 & 900 & 1050& 1200\\
\hline
median $\epsi$ & 7.3e-6 & 0 & 4e-9 & 6e-9\\
max $\epsi$& 2e-4 & 2.8e-5 & 2e-7 & 2.2e-5\\
\end{tabular}
\end{center}

\section{Discussion}
\label{sec:discussion}

\paragraph{Distribution free cluster validation in context} A researcher who wants to discover cluster structure in data must perform several inference tasks. This paper has focused on post-clustering validation, which happens to be the least studied of these inferences. When the data is clusterable, we have shown that validating the clusters is possible, without providing sharp thresholds. In Section \ref{sec:related} we have also cited works that prove that finding the clusters is tractable, under the assumption of clusterability. Hence, the loop is about to close, and we hope that in future work, to integrate the SS method with a clustering algorithm, providing thus a complete ``clustering with guarantees'' methodology.
Furthermore, our distributional results show that for sufficiently
large $n$, the SS method can be the basis of a test for
clusterability, under generic Glivenko-Cantelli assumptions.

\paragraph{Proofs of instability} What happens when the Stability Theorem does not hold for a clustering $\clust$? In this case, the researcher can try to certify that $\clust$ is {\em unstable}. This task is comparatively easier, since a single counterexample $\clust'$ with $\loss(\clust')\approx\loss(\clust)$ and $\distance(\clust',\clust)>\pmin$ suffices. This, again, is a well studied area. The works of \cite{} which propose random perturbations to the data and algorithm are a source of counterexamples. Note that these randomized methods may succeed in proving instability, but they cannot provably guarantee stability without strong additional assumptions on the data. As an alternative worth exploration, one could use the output of the SS method to find a {\em witness} of instability. More precisely, when the SS method fails to produce a valid OI, the matrix $X'$ with $\loss(\dataset,X')\leq \loss(\dataset,\clust)$ is far from $X(\clust)$ in Frobenius norm. One could try to find a clustering $\clust'$ by e.g. rounding $X'$, which would not differ much from $X'$ in either \loss~or distance to $\clust$.

\mmp{ First, by choosing a loss function, she committs herself to a clustering paradigm, that is to the types of clusterings that are favored/detected by the chosen \loss. From here on, the researcher has to (1) use an algorithm to cluster the data, (2) select $K$ (more about this below), (3) decide whether the selected clustering $\clust$ is really meaningful for the data, and sometimes (4) make inferences about the population from which the data was sampled.
 Then, an algorithm is chosen to minimize \loss. The algorithm requires the number of clusters $K$ as input. If, as it is usually the case, $K$ is not known, the data is clusterable}.

\paragraph{Stability and the choice of $K$}
Throughout the paper, we have assumed that $K$ is fixed. We now remark
that the SS method implicitly solves the problem of selecting $K$, and
even that of selecting $\loss$. Indeed, a clustering $\clust$ that is
found stable, with {\em any} $K$ and {\em any} loss function, is a
``correct'' clustering of the data under our paradigm. In the
beginning, we presented the informed user as selecting the $\loss$;
with the concept of stability, one can take another view, of a user
lucky enough to find a stable clustering while searching over loss
functions and $K$ values. The SS method does not preclude the
existence of more than one stable clustering for a data set
$\dataset$. For example, if clusters are hiearchically nested, it is
possible to find stable clusterings at several levels of the
hierarchy.

In practice, $K$ is not known, and it is {\em chosen after} a set of
clusterings $\clust^{(K)}$, with $K=1,2,\ldots K_{max}$ have been
obtained. With the SS method, one could dispense with the (more or
less ad-hoc) methods for selecting $K$ in loss-based
clustering. Indeed, by our initial argument, if 
$\clust^{(K)}$ is proved to be stable for some $K$, this implicitly validates $K$ itself, as well as the loss function used. It is also possible to
select more than one $K$, when the data supports meaningful
partitions with different numbers of clusters.

\paragraph{Silhouette and other cluster quality indices}
%\label{sec:related-silh}
We contrast the framework proposed here with the existing literature on {\em internal cluster validation}; see e.g  \cite{halkidiVazirgiannisHennig:chap15,Arbelaitz:internal-validation-13,henningLiao:13} by indices such as the {\em silhouette} \cite{Rousseeuw:87}. As it is well known, these indices {\em are not associated with a clustering paradigm}, whereas the present paper argues for paradigm specific validation, as a way to assure that the same criterion is used to find the clusters and to validate them. These indices could be potentially used as goodness measures, if their relations to specific clustering loss functions became better understood; the works cited above take steps in this direction.

\paragraph{Comparison with VC bounds} It is extremely rare in statistical inference to have worst case error bounds that are relevant in practice. For instance, the well known VC bounds for the 0-1 classification loss (see e.g. \cite{vapnik98}) typically take values above 1 (hence are completely uninformative from a practical standpoiht) and depend on the VC-dimension, a property of the hypotheses class that is usually intractable to compute.

In contrast, with the SS method, the OI $\epsi$ is always informative when it exists. With SDP relaxations, we obtain bounds that are not only informative, they are near 0 in non-trivial situations. To appreciate how far these guarantees can extend, recall than when $\sigma\approx \frac{1}{6}$ of the center separation, two spherical normal densities start to touch -- no region of low density is left between them. Several of the informative, valid bounds in Section \ref{sec:experiments} are obtained near or even above these critical values. Moreover, an optimality interval is a distribution free, worst case bound.  Thus, we believe that the computational demands of the SDP solver are justified by the guarantees offered.

\section*{Acknowledgement}
The author acknowledges support from NSF DMS award 1810975. This work is completed at the Institute for Pure and Applied Mathematics (IPAM). The author thanks a Simons Fellowship from IPAM. Also the author gratefully thanks the Pfaendtner and Tkatchenko labs,  especially Chris Fu and Stefan Chmiela for providing both data and expertise. 

\bibliography{stability}

\appendix
\appendix
\section*{Proofs}
We first state several helpful propositions needed for our proofs.

\begin{prop} \label{prop:normF}
For any $X\in\X$, $||X||_F^2\leq K$. 
\end{prop}

\begin{prop}
For any fixed two clusterings $\C,\C'$ it holds that
\begin{equation}
    |d_{\Pp}^{EM}(\C,\C')-d_{\widehat{\Pp}}^{EM}(\C,\C')|\leq \sqrt{\frac{\log(4/\delta)}{2n}}
\end{equation}
with probability $1-\delta/2$
\label{prop: empdist}
\end{prop}

\paragraph{Proof of Proposition \ref{prop:X}}
$X_{ij}\in[0,1]$ is obvious from the definition \eqref{eq:X}.
\beq
\trace X\,=\,\sum_{k=1}^K\sum_{i\in C_k}X_{ii}\,=\,\sum_{k=1}^K\sum_{i\in C_k}\frac{1}{n_k}\,=\,\sum_{k=1}^Kn_k\frac{1}{n_k}\,=\,K.
\eeq
Denote by $k_0$ the cluster containing data point $i$.
\beq
(X\bfone)_i\,=\,\sum_{j=1}^n X_{ij}\,=\,\sum_{k=1}^K\sum_{i\in C_k}X_{ij}
\sum_{j\in C_{k_0}}\frac{1}{n_k}\,=\,1.
\eeq
Moreover, $X=ZZ^T$, hence $X\succeq 0$.\hfill $\Box$

\paragraph{Proof of Proposition \ref{prop:epsi2}}
Let $X',X'_2$ be  optima for the SS, respectively SS2 problem.  Then, by the triangle inequality,
\beqa
\epsi'\;=\;||X'-X(\clust)||&\leq&||X^*-X(\clust)||+||X^*-X'||\\
&\leq&||X^*-X'_2||+||X^*-X'_2||\;=\;2\epsi'2. \hfill\Box
\eeqa

\paragraph{Proof of Theorem \ref{prop:mebound}}
Note that for any clustering $\clust$, $||X(\clust)||_F^2=\sum_{i,j=1}^nX_{ij}^2=,=\sum_{k=1}^Kn_k^2\left(\frac{1}{n_k}\right)^2=K$. Moreover, from proposition \ref{prop:normF} we have $||X||_F^2\leq K$.

Note also that $||X-X'||^2_F=||X||^2_F+||X'||^2_F-2\langle X,X'\rangle=2K-2\langle X,X'\rangle$. Hence, the optimization problem $\probe$ finds the feasible $X'$ which is furthest away from $X$. 
\comment{{\bf Proof sketch} We frame this result in a Proposition because of its importance rather than because of its technical difficulty (the complete proof is in the supplement). Since any clustering $\clust'$ that is better than $\clust$ in terms of \lossk~must be feasible for $\probe$, $\langle X(\clust),X(\clust')\rangle \geq \delta$. That is, the two clusterings cannot be too ``disaligned'' w.r.t the Frobenius scalar product. Because $||X(\clust')||^2_F=K$ for any clustering, it is also easy to see that $\langle X(\clust),X(\clust')\rangle=K-\frac{1}{2}||X(\clust)-X(\clust')||^2_F$.} This completes Step 1. For Step 2 we can apply Theorem 9 of \cite{M:equivalence-ML10}, which bounds the earthmover distance $\distance$. \hfill $\Box$

% whenever $\delta=K-||Z(\clust)^TZ(\clust')||^2F\leq \pmin(\clust)$, $d(\clust,\clust')\leq \delta\pmax(\clust)$.

\paragraph{Proof of Proposition \ref{prop:normF}} Denote by $\lambda_{1},\ldots \lambda_n$ the eigenvalues of $X$. 
Since $X$ has non-negative elements, and $X\bfone=\bfone$, by the Frobenius Theorem, $|\lambda_i|\leq 1$, and because $X\succeq 0$, $\lambda_i\geq 0$ for all $i\in [n]$. Hence, $\lambda_i\in[0,1]$, for all $i$, and $||X||_F^2=\trace X^2=\sum_{i=1}^n \lambda_i^2\leq \sum_{i=1}\lambda_i=\trace X=K$.\hfill $\Box$

\paragraph{Proof of Theorem \ref{prop:ncut}}
The proof is similar to the proof of Proposition \ref{prop:mebound}, after noting that for any $\clust$, $||X(\clust)||^2_F=K$. Again we use Theorem 9 of \cite{M:equivalence-ML10} which relates the error in Frobenius norm now to weigted EM distance. Note that $\distance_{w}$ can always bound $\distance$ by a factor of $\max_i w_i/\min_i w_i$. 

\paragraph{Proof of Theorem \ref{prop:generic-XZ}}
For any convex problem of the form \eqref{eq:convexrel}, adding the
constraint $\loss\leq l$ and a linear objective preserves
convexity. The functions $\langle X,X'\rangle,\langle Z,Z'\rangle,\langle\tilX,\tilX'\rangle$ are obviously
linear in the second variable. Hence, the SS problem is always
convex. Moreover, if \eqref{eq:convexrel} has a non-empty relative
interior and $X\neq X^*$, the SS problem also has a non-empty
relative interior, hence strong duality holds. Same arguments hold for $\tilX,Z$.  Now, for $X$, is is
easy to see from section \ref{sec:kmeans} that the proof Proposition
\ref{prop:mebound} holds regardless of the space $\X$ or of the
expression of \loss. 

For $Z$,  we first notice that $X=ZZ^T$ hence we can prove the result if we can lower bound $||Z^TZ'||^2_F$ for any pair of clusterings. We have $\langle Z,Z'\rangle=\trace Z^TZ'$. Now, for any symmetric matrix $A$ with non-negative elements, $||A||^2_F=\trace A^2=\sum_{i\in [n]}\lambda_i(A)^2\geq\frac{1}{2}(\sum_{i\in[n]}\lambda_i(A))^2=\frac{1}{2}(\trace A)^2$. Let $\delta'=\delta^2/2$. Then $\epsi=(K-\delta')\pmax$ is an OI whenever it is smaller or equal to $\pmin$, by an argument similar to the proof of Proposition \ref{prop:mebound}. Alternatively, we can notice that $||Z^TZ'||_F^2=\trace (Z^TZ')^T(Z^TZ') 
=\trace (ZZ^T)(Z'(Z')^T)=\trace XX'=\langle X,X'\rangle$.

For the $\tilX$ representation, we note that $||\tilX||_F^2=\sum_{k\in[K]}n_k^2\leq (n-K+1)^2+(K-1)$ for any $\tilX$ representing a clustering. Hence, $||\tilX-\tilX'||^2_F=||\tilX||^2_F+||\tilX'||^2_F-2\langle \tilX,\tilX'\rangle\leq ||\tilX||^2_F+(n-K+1)^2+(K-1)-2\delta$. We now apply Theorem 27 of \cite{M:equivalence-ML10} which states that $\distance(\clust,\clust')\leq \frac{1}{2n^2\pmin}||\tilX(\clust)-\tilX(\clust')||^2_F$ and obtain $\epsi=\frac{\sum_{k\in[K]}n_k^2+(n-K+1)^2+(K-1)-2\delta}{2\pmin}$ whenever $\epsi\leq \pmin$.
\hfill$\Box$

\paragraph{Proof of Theorem \ref{thm:instable}}
On the sample with probability $1-\delta/2$ it holds that 
\begin{equation}
    \sup_{\C\in \mathbf{C}_K(\dataset)} |L(\C)-\widehat{L}(\C)|\leq \Psi(n,\frac{\delta}{2}).
    \label{uniformbound}
\end{equation}
Now we condition on the event that  \eqref{uniformbound} holds. 

If there exists a population minimizer $\C^{opt}$ such that $d_{\Pp}^{EM}(\C^{opt},\widehat{\C}^{opt})\leq \epsilon_0/2$. Note that by assumption of instability, there exists another clustering $\C^{*}$ such that $d_{\Pp}^{EM}(\C^*,\C^{opt})>\epsilon_0$ and $L(\C^{*}) < L(\C^{opt})+\eta$, then $d_{\Pp}^{EM}(\C^{*},\widehat{\C}^{opt})>\epsilon_0/2$. We can bound
\begin{equation}
    \widehat{L}(\C^{*})\leq L(\C^{*})+\Psi(n,\frac{\delta}{2}) \leq L(\C^{opt}) + \Psi(n,\frac{\delta}{2}) + \eta \leq L(\widehat{C}^{opt}) +  \Psi(n,\frac{\delta}{2}) + \eta \leq \widehat{L}(\widehat{\C}) + 2\Psi(n,\frac{\delta}{2}) + \eta
\end{equation}
and take $\widehat{\C}'=\C^{*}$.

If such a clustering $\C^{opt}$ does not exist, then note that for any optimal solution of the population clustering $\C^{opt}$,
\begin{equation}
    \widehat{L}(\C^{opt})\leq L(\C^{opt})+\Psi(n,\frac{\delta}{2}) \leq L(\widehat{\C})+\Psi(n,\frac{\delta}{2}) \leq \widehat{L}(\widehat{\C})+2\Psi(n,\frac{\delta}{2})
\end{equation}
and take $\widehat{\C}'=\C^{opt}$.

In both cases, applying proposition \ref{prop: empdist} it holds that $d_{\Pp}^{EM}(\widehat{\C}',\widehat{\C}^{opt})\geq \epsilon_0/2-\sqrt{\log(4/\delta)/2n}$. Therefore $\widehat{\C}^{opt}$ is $(\Delta + 2\Psi(n,\frac{\delta}{2}), \epsilon_0/2-\sqrt{\log(4/\delta)/2n})$ instable.
\hfill$\Box$

\paragraph{Proof of Proposition \ref{prop: empdist}}
For fixed permutation $\pi$, from Hoeffding's inequality
    \begin{equation}
        \left|\mathbb{E}_{\Pp} \sum_{i=1}^K\bm{1}_{X\in C_i\cap C_{\pi(i)}}-\mathbb{E}_{\widehat{\Pp}} \sum_{i=1}^K\bm{1}_{X\in C_i\cap C_{\pi(i)}}\right|\leq \sqrt{\frac{\log(2/\delta)}{2n}}
    \end{equation}
    with probability $1-\delta$.
    Now let $\pi^*,\widehat{\pi}^*$ be the permutation maximizing $\mathbb{E}_{\Pp} \sum_{i=1}^K\bm{1}_{X\in C_i\cap C_{\pi(i)}}$ and $\mathbb{E}_{\widehat{\Pp}} \sum_{i=1}^K\bm{1}_{X\in C_i\cap C_{\pi(i)}}$ respectively. Then 
    \begin{align}
        \mathbb{E}_{\Pp} \sum_{i=1}^K\bm{1}_{X\in C_i\cap C_{\pi^*(i)}}\leq \mathbb{E}_{\widehat{\Pp}} \sum_{i=1}^K\bm{1}_{X\in C_i\cap C_{\pi^*(i)}}+\sqrt{\frac{\log(2/\delta)}{2n}}\leq \mathbb{E}_{\widehat{\Pp}} \sum_{i=1}^K\bm{1}_{X\in C_i\cap C_{\widehat{\pi}^*(i)}}+\sqrt{\frac{\log(2/\delta)}{2n}}\\
        \mathbb{E}_{\widehat{\Pp}} \sum_{i=1}^K\bm{1}_{X\in C_i\cap C_{\widehat{\pi}^*(i)}}\leq \mathbb{E}_{\Pp} \sum_{i=1}^K\bm{1}_{X\in C_i\cap C_{\widehat{\pi}^*(i)}}+\sqrt{\frac{\log(2/\delta)}{2n}}\leq \mathbb{E}_{\Pp} \sum_{i=1}^K\bm{1}_{X\in C_i\cap C_{\pi^*(i)}}+\sqrt{\frac{\log(2/\delta)}{2n}}
    \end{align}
    Therefore one concludes that $|d_{\Pp}^{EM}(\C,\C')-d_{\widehat{\Pp}}^{EM}(\C,\C')|\leq \sqrt{\log(2/\delta)/2n}$ with probability $1-\delta$.
\hfill$\Box$

 %%% removed additional experiments. Go to
                                 % nips paper supplement for them.

\end{document}